\documentclass[11pt]{article}

\usepackage[final]{acl}

\usepackage{times}
\usepackage{latexsym}

\usepackage[T1]{fontenc}

\usepackage[utf8]{inputenc}

\usepackage{microtype}

\usepackage{inconsolata}

\usepackage{graphicx}

\usepackage{amsmath,amssymb}
\usepackage{todonotes} 
\usepackage{lineno}
\usepackage{subcaption}
\usepackage{listings}
\usepackage[most]{tcolorbox}
\tcbuselibrary{listings,breakable}

\newcommand{\info}[1]{}
\newcommand{\fix}[1]{#1}

\newcommand{\infon}[1]{}
\newcommand{\fixn}[1]{#1}
\tcbuselibrary{listings}
\usepackage{listings}
\usepackage[table]{xcolor}

\definecolor{methodgreen}{RGB}{230,245,230}
\definecolor{promptbackground}{RGB}{248,248,248}
\definecolor{promptborder}{RGB}{170,170,170}
\definecolor{prompttitle}{RGB}{240,240,240}

\newtcblisting{promptbox}[1]{
    enhanced,
    listing only,
    title={#1},
    width=\linewidth,
    fonttitle=\bfseries\footnotesize,
    coltitle=black,
    colbacktitle=prompttitle,
    colback=promptbackground,
    colframe=promptborder,
    boxrule=0.4pt,
    arc=1mm,
    outer arc=1mm,
    boxsep=0pt,
    left=1.5mm,
    right=1.5mm,
    top=1.2mm,
    bottom=1.2mm,
    toptitle=0.8mm,
    bottomtitle=0.8mm,
    before skip=0pt,
    after skip=0pt,
    listing options={
        basicstyle=\ttfamily\scriptsize,
        breaklines=true,
        breakatwhitespace=true,
        columns=fullflexible,
        keepspaces=true,
        showstringspaces=false,
        tabsize=2,
        aboveskip=0pt,
        belowskip=0pt
    }
}

\newtcblisting{routepromptbox}[1]{
    enhanced,
    listing only,
    title={#1},
    width=\linewidth,
    fonttitle=\bfseries\footnotesize,
    coltitle=black,
    colbacktitle=prompttitle,
    colback=promptbackground,
    colframe=promptborder,
    boxrule=0.4pt,
    arc=1mm,
    outer arc=1mm,
    boxsep=0pt,
    left=1.5mm,
    right=1.5mm,
    top=1.2mm,
    bottom=1.2mm,
    toptitle=0.8mm,
    bottomtitle=0.8mm,
    before skip=0pt,
    after skip=0pt,
    listing options={
        basicstyle=\ttfamily\scriptsize,
        breaklines=true,
        breakatwhitespace=true,
        columns=fullflexible,
        keepspaces=true,
        showstringspaces=false,
        tabsize=2,
        aboveskip=0pt,
        belowskip=0pt
    }
}

\title{Instructions for *ACL Proceedings}

\author{
 {
  \textbf{Yuxiang Wang}, 
  \textbf{Junhao Gan}, 
  \textbf{Jianzhong Qi} 
  }
  \\
  {The University of Melbourne} 
  \\ \texttt{yuxiang.wang8@student.unimelb.edu.au}\\ 
   \texttt{\{junhao.gan, jianzhong.qi\}@unimelb.edu.au}
}

\usepackage{times}
\usepackage{latexsym}
\usepackage[T1]{fontenc}
\usepackage[utf8]{inputenc}
\usepackage{microtype}
\usepackage{inconsolata}
\usepackage{amsmath,amssymb}
\usepackage{booktabs}
\usepackage{multirow}
\usepackage{enumitem}
\usepackage{graphicx}
\usepackage{xcolor}
\usepackage{array}
\usepackage{makecell}
\usepackage{hyperref}
\usepackage{booktabs}
\usepackage{multirow}
\usepackage{graphicx}
\usepackage[table]{xcolor}
\usepackage{pifont}
\newcommand{\cmark}{\ding{51}}
\newcommand{\xmark}{\ding{55}}

\newcommand{\wdata}{\texttt{WikiTQ}}
\newcommand{\bdata}{\texttt{BirdQA}}
\newcommand{\ldata}{\texttt{WTQ-L}}
\newcommand{\sdata}{\texttt{SLQA}}
\newcommand{\tdata}{\texttt{TableBench}}

\newcommand{\method}{\textsc{TableRoute}}
\newcommand{\readroute}{\textsc{Direct}}
\newcommand{\reduceroute}{\textsc{Reduce}}
\newcommand{\execroute}{\textsc{Execute}}

\newcommand{\best}[1]{\textbf{#1}}
\newcommand{\second}[1]{\underline{#1}}

\title{\method{}: Capability-Aware Routing for Scalable Table Question Answering}

\begin{document}
\maketitle

\begin{abstract}
Table question answering with large language models remains difficult to scale: tables range from compact to million-token sizes, while questions require precise grounding of columns, values, and computations. 
Existing methods often specialize in a fixed inference strategy, which is brittle across different table scales and model capabilities. 
Full-table reading is efficient for compact inputs but suffers from irrelevant context. Table reduction helps long inputs but depends on reliable evidence localization, while execution is required when tables exceed the model context window and remains sensitive to schema and value grounding errors.
We propose \method{}, a capability-aware routing framework with evidence-conditioned inference for scalable TableQA. 
\method{} first generates validated question--table evidence and identifies the required operations, then routes each instance among these three strategies
\info{R1-Q1/2, R2-Q1/4, R3-Q1}
\fix{according to serialized table length and calibrated estimates of how reliably the base model handles these operations.}
The validated evidence supports the selected route by grounding columns, values, and conditions.
\infon{AC-Q1, AC-Q2, R1-W1, R1-W2}\fixn{Across \wdata, \ldata, \bdata, and \tdata, \method{} consistently improves accuracy through complementary routing and evidence conditioning across table scales, while reducing online token cost.} Code will be made available upon publication.
\end{abstract}

\section{Introduction}

Table question answering (TableQA) requires answering natural language questions grounded in tables~\cite{pasupat}. Recent TableQA systems increasingly use LLMs as backbones, but the task remains challenging because tables encode information through both cell contents and schema, requiring schema--value alignment, evidence localization, cell normalization, and exact computation~\cite{cao2026tablemaster, wang2025accurate}.

\begin{figure}[t]
\centering
\includegraphics[width=\columnwidth]{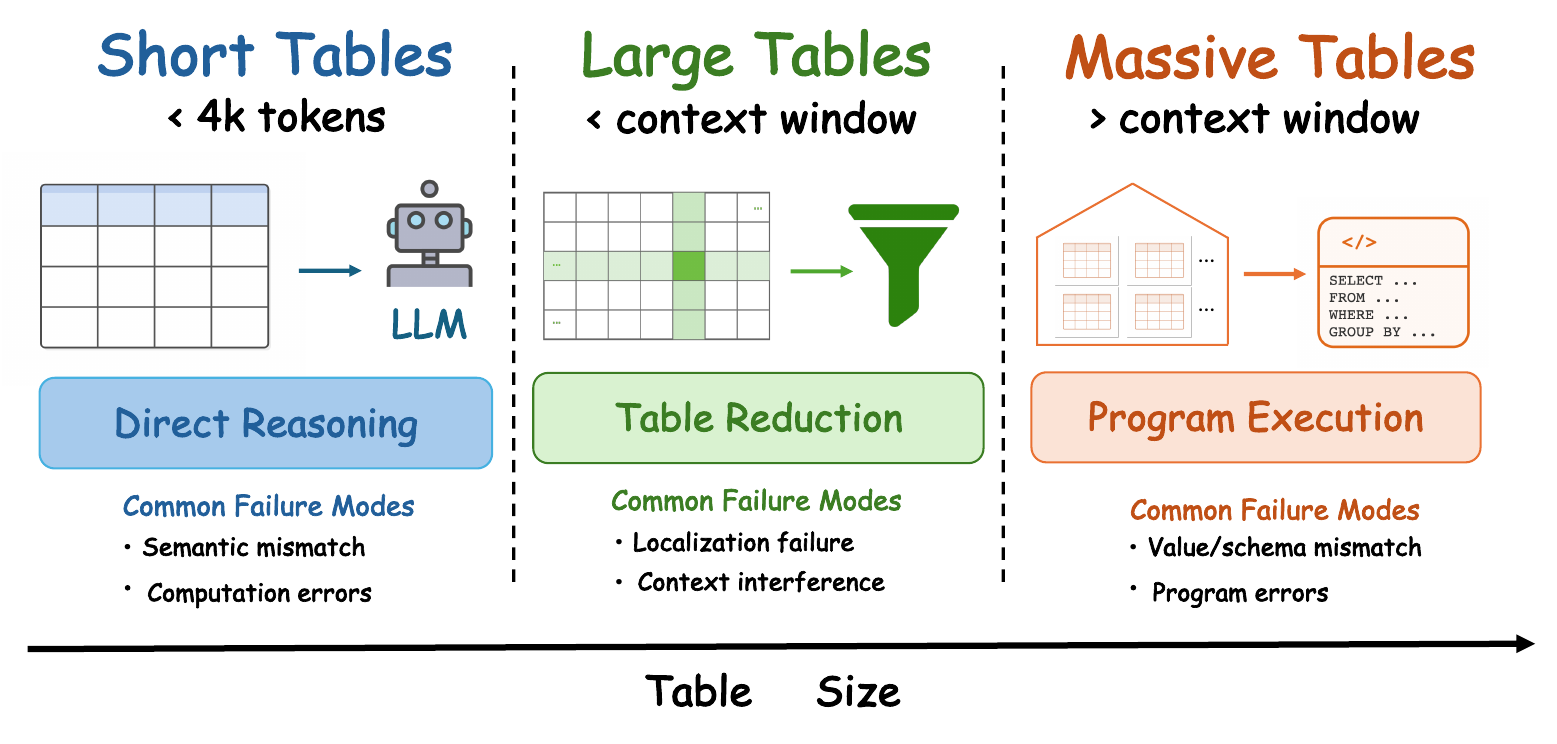}
\caption{Motivation for scalable TableQA.}
\label{fig:intro-motivation}
\end{figure}

Despite recent progress, TableQA remains difficult in practical settings where table size can range from compact inputs to million-token scale. Our analysis shows that semantic mismatch and computation errors persist across scales: models may map questions to wrong columns or values, and textual reasoning often fails on exact numerical operations. As tables grow longer, relevant evidence becomes harder to locate among irrelevant rows. When tables exceed the model context window, \readroute\ reading becomes infeasible and program \execroute\ is required, but generated programs can still fail due to incorrect schema grounding, value matching, or operation parameters. These failures echo recent analyses of TableQA reasoning and long-table \reduceroute~\citep{syntqa,cao2026tablemaster,talon,tadre}. 
\info{R1-Q1/2, R2-Q1/4}
\fix{
Together, these challenges motivate three complementary, question-dependent inference strategies for scalable TableQA: \readroute\ answering, table \reduceroute, and program \execroute. Motivated by model routing, which matches queries to model capabilities~\citep{routellm,irt}, a scalable system should consider table size, the base model's direct-reasoning capability,  and the operations required by the question. The same input length may remain manageable for lookup-style reasoning but become unreliable for aggregation or arithmetic. Moreover, every strategy requires reliable grounding to table columns, values, and conditions. Figure~\ref{fig:intro-motivation} summarizes this motivation.}

\info{R1-Q1/2/4, R2-Q1/4, R3-Q1}
\fix{
We propose \method{}, a capability-aware framework combining validated evidence with single-route selection.
Given a question, table, and base model, \method{} grounds question spans to table columns, values, and conditions, and identifies required operations.
It compares the table length against a calibrated reliable range for the base model and required operations.
Inputs within this range use \readroute{}, longer context-fit inputs use \reduceroute{}, and over-window inputs use \execroute{}. The selected route reuses the validated evidence to ground direct reasoning, guide operation-based reduction, or constrain program generation.
}

\noindent\mbox{\textbf{Our Contributions.}}
\info{R1-Q1/2/4, R2-Q1/4, R3-Q1}
\fix{
\underline{{(1) Capability-Aware Routing.}}
We formulate scalable TableQA as per-question selection among \readroute, \reduceroute, and \execroute\ using reliable table-size ranges calibrated to base-model capability and operations.
\infon{AC-Q1, R1-W1, R1-W3}\fixn{\underline{{(2) Evidence-Conditioned Inference.}} We introduce a shared evidence layer that identifies required operations and provides structurally checked column, value, and condition grounding for route-specific inference. \underline{{(3) Across-Scale Improvements.}} Experiments across table scales and backbones demonstrate accuracy gains from routing and evidence conditioning, with lower online token cost.}}

\section{Related Work}
\label{related}

\paragraph{Direct reasoning for TableQA.}
A common approach to LLM-based TableQA is to provide the serialized full table as context and answer directly from the question--table prompt.
This paradigm can be enhanced through chain-of-thought prompting in TaCo~\cite{COT,ZhengYJLLSW23}, row-wise traversal in RoT~\citep{rot}. 
While effective when relevant evidence is accessible, direct reasoning becomes less reliable on long tables due to irrelevant context and lost-in-the-middle effects~\citep{LiuLHPBPL24}.
\method{} limits direct reasoning to reliable table-size ranges calibrated for each model and operation.

\paragraph{Table reduction for TableQA.}
Reduction-based methods construct question-relevant contexts before answering. Binder decomposes questions into executable SQL sub-programs~\citep{binder}; Dater decomposes evidence and questions before answer filling~\citep{dater}; Chain-of-Table transforms tables during reasoning~\citep{chainoftable}; TabSQLify uses Text-to-SQL to select sub-tables~\citep{TabSQLify}; TabScope reduces tables through operation-aware decomposition~\citep{tabscope}.
Recent methods improve evidence selection through graph reranking, graph traversal, entity-oriented retrieval, or structured decomposition and verification~\citep{tabgr,graphotter,TUNES,TIDE}.
While better suited to long tables, these methods may discard evidence needed for multi-step reasoning or introduce invalid operations through incorrect localization and decomposition.
Our \reduceroute{} instead generates and verifies a compact operation list grounded in validated columns, values, and conditions.

\paragraph{Program execution for TableQA.}
Program-based methods answer questions through executable operations. Text-to-SQL methods improve SQL generation through schema linking, decomposition, verification, or multi-agent orchestration, including TQ-SQL, DAIL-SQL, MAC-SQL, and DeepEye-SQL~\citep{tqsql,DAILSQL,macsql,li2025deepeye}. H-STAR further combines SQL execution with textual reasoning~\citep{hstar}. Python-based methods such as TableRAG and TALON use generated programs or tool calls to explore and compute over large tables~\citep{tabrag,talon}. Execution is necessary when the complete input exceeds the model context window, but remains sensitive to value mismatch, incorrect column grounding, and operation errors.
\method{} conditions the execution route on validated columns, values, and conditions to constrain schema linking, value matching, and program generation.

\begin{figure*}[t]
    \centering
    \begin{minipage}[t]{0.49\textwidth}
        \centering
        \includegraphics[height=1.5in,keepaspectratio]{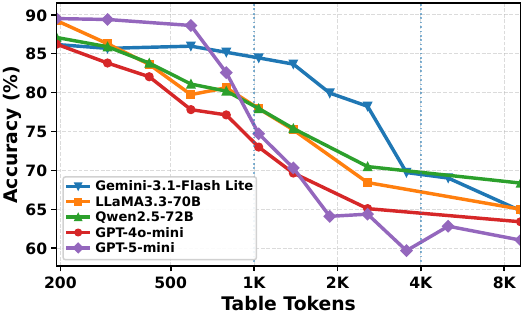}\\[-0.5ex]
        \small (a) Full-table QA accuracy vs. Table tokens.
    \end{minipage}
    \hfill
    \begin{minipage}[t]{0.49\textwidth}
        \centering
        \includegraphics[height=1.5in,keepaspectratio]{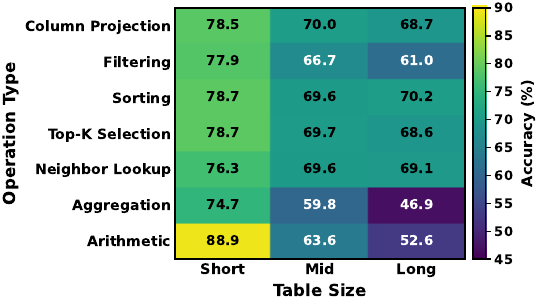}\\[-0.5ex]
        \small (b) Accuracy by operation types and table length.
    \end{minipage}
    \caption{
    Full-table QA accuracy as table size increases. 
    In (a), diagnostic models are re-binned into  token ranges. 
    In (b), questions are grouped by operator category (definitions in Appendix~\ref{app:operator-definitions}), with results pooled across models.
    }
    \label{fig:size-effects}
\end{figure*}

\paragraph{Adaptive routing for TableQA.}
The above strategies are complementary: full-table answering suits compact inputs, reduction handles long context-fit tables, and execution is necessary for over-window tables.
Recent systems combine them through LLM self-routing, post-hoc answer selection, adaptive textual--symbolic reasoning, or multi-agent collaboration~\citep{syntqa,cao2026tablemaster,tablecritic,datafactory}.
Self-routing is inexpensive but may misjudge route suitability, while answer selection and multi-agent methods improve robustness by running multiple paths at higher cost, as analyzed in Section~\ref{sec:challenge-analysis}.
\info{R1-Q1/2/6, R2-Q1/4/5, R3-Q1/2}
\fix{
\method{} instead calibrates route reliable range and inferences using single route with validated evidence.
}
Table~\ref{tab:method_comparison} in Appendix~\ref{app:method_comparison} compares their routing basis, evidence use, executed paths, and inference cost.

\section{Empirical Analysis of Scalable TableQA}
\label{sec:challenge-analysis}

Before introducing \method{}, we examine how TableQA behavior changes across table scales and inference strategies.
We group examples into \textit{short tables} with fewer than $4$K table tokens, \textit{long tables} that remain within the model context window, and \textit{massive tables} whose inputs exceed the context window. We ask three questions: which errors dominate across scales, how table size and required operations affect full-table answering, and whether existing route-switching strategies effectively exploit route complementarity. Additional setup and results are provided in Appendix~\ref{app:challenge-analysis}.

\subsection{RQ1: What errors dominate TableQA across table scales?}
\label{sec:rq1-error-types}

\paragraph{Setup.}
We manually analyze 200 randomly sampled incorrect cases for each table scale and assign the first responsible failure to one of five categories: semantic mismatch, computation error, evidence localization, reduction or execution error, and others. 
We include representative direct reasoning, table reduction, and program execution baselines. Detailed definitions and annotation procedures are provided in Appendix~\ref{app:challenge-analysis}.

\begin{table}[t]
\centering
\scriptsize
\resizebox{\columnwidth}{!}{
\begin{tabular}{lcccc}
\toprule
Error Type & Short & Long & Massive & Overall \\
\midrule
Semantic mismatch & 44.0 & 31.0 & 52.0 & 42.3 \\
Computation error & 24.0 & 28.0 & -- & 17.4 \\
Evidence localization & 17.0 & 26.0 & -- & 14.3 \\
Reduction or execution error & 7.0 & 11.0 & 41.0 & 19.7 \\
Others & 8.0 & 4.0 & 7.0 & 6.3 \\
\midrule
Semantic + computation & 68.0 & 59.0 & 52.0 & 59.7 \\
\bottomrule
\end{tabular}
}
\caption{Error distributions (\%) over 200 sampled failures per scale. For massive tables, evidence localization and computation failures are counted as reduction or execution errors because the answer generation step is programmatic.}
\label{tab:error-distribution}
\end{table}

\info{R1-Q4, R2-Q1/2}
\fix{
\paragraph{Results.}
Table~\ref{tab:error-distribution} shows that semantic mismatch is the largest error category, accounting for 42.3\% of failures overall and remaining prominent at every scale.
On long tables, computation and evidence localization errors account for 28.0\% and 26.0\% of failures, respectively.
On massive tables, reduction or execution errors rise to 41.0\%.
Representative Text-to-SQL results in Appendix~\ref{app:massive-execution-baselines} further show that execution remains sensitive to schema and value grounding.
These results indicate that switching inference strategies does not eliminate semantic grounding errors, and each route must therefore preserve explicit mappings from question spans to table columns and cell values, combining with the corresponding conditions.
}

\info{R1-Q2, R2-Q4/5}
\fix{
\subsection{RQ2: How do table sizes and operation types affect full-table answering?}
}
\label{sec:rq2-size}

\paragraph{Setup.}
We analyze full-table answering on \wdata{} using multiple diagnostic LLMs. 
For each model, we group examples by table tokens and report accuracy trends in Figure~\ref{fig:size-effects}(a). 
To test whether this length effect depends on the operations required by questions, we further group examples by operation types and report accuracy across table-length bins in Figure~\ref{fig:size-effects}(b). 

\begin{figure*}[t]
    \centering
    \includegraphics[width=1\textwidth]{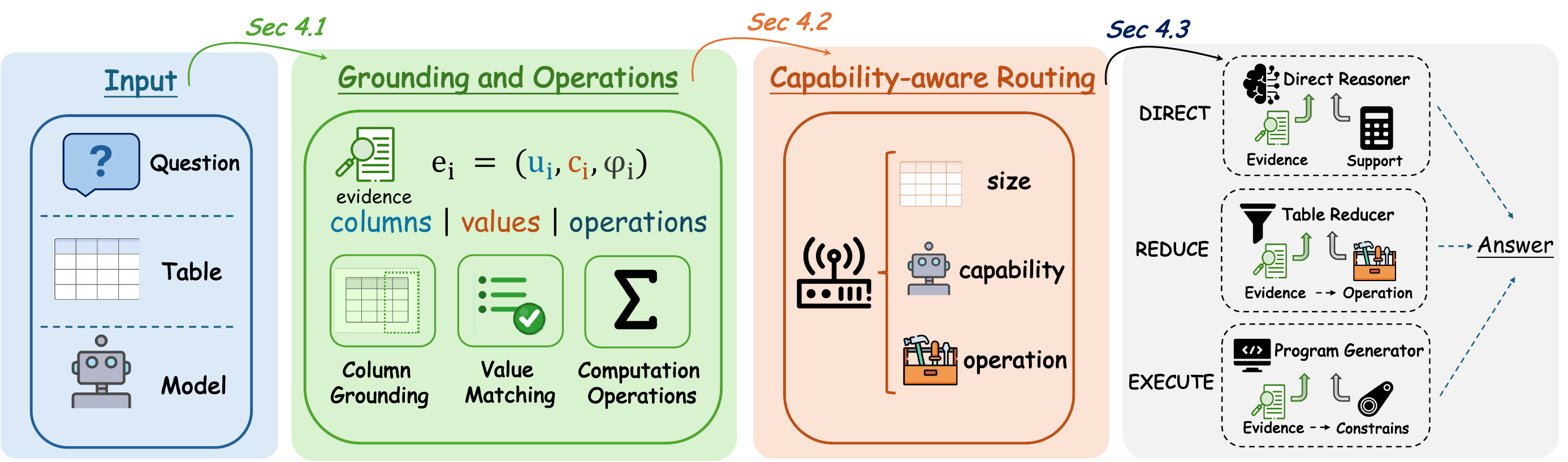}
    \caption{
   Overview of \method{}.
\method{} generates validated evidence and required operations, performs
capability-aware routing, and reuses the evidence for \textsc{Direct}, \textsc{Reduce}, or \textsc{Execute} inference.
    }
    \label{fig:method-overview}
\end{figure*}

\paragraph{Results.}
\info{R1-Q1/2, R2-Q4/5, R3-Q1}
\fix{
Figure~\ref{fig:size-effects}(a) shows that full-table accuracy generally decreases as table size increases, but different base models begin to degrade at different table sizes and decline at different rates. Figure~\ref{fig:size-effects}(b) further shows that this effect varies across required operations. 
Aggregation and arithmetic exhibit the steepest declines, while the remaining categories show smaller or less uniform changes across table-size ranges. Table size is therefore a strong indicator of full-table reliability, but a single length-only profile cannot describe every model and operation category. The reliable table-size range of full-table answering depends on both the base model and the operations required by the question.}

\subsection{RQ3: How reliably do existing strategies switch among TableQA routes?}
\label{sec:rq3-hybrid}
\paragraph{Setup.}
\info{R3-Q1}
\fix{
The preceding analysis shows that no single inference route is suitable across all table scales and questions.
We compare two existing switching strategies.
\textit{LLM self-routing} asks the model to select a route before answering, while \textit{answer selection} executes multiple routes and chooses among their outputs~\citep{cao2026tablemaster,syntqa}.
We use GPT-4o-mini as the common base model and report both accuracy and relative token cost. Oracle routing is included only as an analysis upper bound that selects the correct output whenever any candidate route succeeds.
}

\begin{table}[t]
\centering
\resizebox{\linewidth}{!}{%
\scriptsize
\setlength{\tabcolsep}{4pt}
\begin{tabular*}{\columnwidth}{@{\extracolsep{\fill}}lcccc@{}}
\toprule
\multirow{2}{*}{Strategy}
& \multicolumn{2}{c}{\ldata}
& \multicolumn{2}{c}{\bdata} \\
\cmidrule(lr){2-3}
\cmidrule(lr){4-5}
& Acc. & Relative Cost ($\times$) & Acc. & Relative Cost ($\times$) \\
\midrule
CoT only
& 58.1 & 1.00 & -- & -- \\
DATER only
& 54.7 & 3.15 & 61.8 & 2.76 \\
DeepEye-SQL only
& -- & -- & 69.7 & 1.00 \\
\midrule
LLM Self-routing
& 60.4 & 1.48 & 66.8 & 1.34 \\
Answer Selection
& 62.0 & 4.15 & 69.0 & 3.76 \\
Oracle Routing
& 69.1 & -- & 77.6 & -- \\
\bottomrule
\end{tabular*}
}
\caption{
Comparison of route-switching strategies.
CoT, DATER, and DeepEye-SQL represent the \readroute, \reduceroute, and \execroute\ routes, respectively.
Oracle Routing is an analysis-only upper bound.
Cost is reported relative to the corresponding single-route baseline.
}
\label{tab:hybrid-main}
\end{table}

\paragraph{Results.}
\info{R1-Q1, R2-Q5, R3-Q1}
\fix{
Table~\ref{tab:hybrid-main} confirms route complementarity. Oracle routing exceeds the strongest single route by $11.0$ points on \ldata{} and $7.9$ points on \bdata{}.
However, self-routing remains substantially below the oracle, indicating that model self-assessment does not reliably identify the best route.
Answer selection narrows the accuracy gap but raises token cost to $4.15\times$ on \ldata{} and $3.76\times$ on \bdata{} by executing multiple routes.
These results show that route complementarity is substantial, but existing switching strategies either estimate route suitability unreliably or incur the cost of multi-route inference.
}

\section{Method}
\label{sec:method}

We propose \method{}, a capability-aware framework for single-route TableQA. Given a question $q$, table $T$, and base model $M$, it generates validated question--table evidence and identifies the required operations.
It then selects one route based on the input table length and the estimated reliability of full-table answering, and reuses the validated evidence for route-specific inference. Figure~\ref{fig:method-overview} presents the overall framework.

\subsection{Question Grounding and Operations}
\label{sec:evidence}

\info{R1-Q4, R2-Q1/2/3}
\fix{
Before routing, \method{} produces a validated evidence set
$E(q,T)=\{e_i\}_{i=1}^{m(q,T)}$
and a required-operation set $\mathcal{O}(q,T)$ per question--table pair.
Here, $m(q,T)$ is the number of retained evidence items, allowing multiple grounding relations per question.
The evidence supports the selected route, while the required operations inform routing.
}

\paragraph{Evidence Generation.}
\method{} first generates candidate question--table evidence and required operations in a structured format, then validates them against $T$.
Each retained evidence item is represented as
\vspace{-0.7cm}
\begin{center}
\small
\begin{equation}
    e_i=(u_i,c_i,\phi_i),
    \label{eq:evidence-item}
\end{equation}
\end{center}
where $u_i$ is the question span triggering grounding, $c_i$ is the resolved table column, and $\phi_i$ is the condition over $c_i$, specifying how a grounded value is matched or compared.
If no condition is required, $\phi_i=\varnothing$.
\infon{AC-Q2, R1-W3}\fixn{Validation checks column resolution, type compatibility, value parsing, and condition executability, removing invalid candidates without computing answers. Throughout this paper, validated evidence refers to evidence that passes these structural and executability checks.} If no evidence is retained, the selected route proceeds with its standard input.
Prompts and validation details are provided in Appendix~\ref{app:evidence-generation}.

\paragraph{Operation Grouping.}
\info{R1-Q1/2/3/4, R2-Q2/4/5, R3-Q1/3}
\fix{
The required-operation set $\mathcal{O}(q,T)$ covers seven fine-grained types.
To avoid sparse per-type calibration subsets, we group them by their burden on direct full-table reasoning.
The retrieval-oriented \textsc{Light} group contains
\textsc{Projection}, \textsc{Filter}, and \textsc{Neighbor}, while the
reasoning-intensive \textsc{Heavy} group contains
\textsc{Sort}, \textsc{TopK}, \textsc{Aggregate}, and \textsc{Compute}.
We denote the assigned group by
$z(q,T)\in\{\textsc{Light},\textsc{Heavy}\}$, where any question requiring a \textsc{Heavy} operation is assigned to \textsc{Heavy}.
The grouping is used only for capability calibration, while the original operations remain available to the selected route when needed.
}

\info{R1-Q4, R2-Q2}
\fix{
For example, in ``How many Australian teams scored above 10?'',
the evidence grounds \textit{Australian} as
$\texttt{Country}=\texttt{Australia}$ and \textit{above 10} as
$\texttt{Score}>10$.
$\mathcal{O}(q,T)=
\{\textsc{Filter},\textsc{Aggregate}\}$.
Since it requires aggregation, it is assigned
$z(q,T)=\textsc{Heavy}$.
}

\subsection{Capability-Aware Routing}
\label{sec:routing}

\info{R1-Q1/2, R2-Q1/4/5, R3-Q1}
\fix{
Since full-table reliability depends on table size, base model, and required operations, we calibrate reliable size ranges for each model--operation group using held-out full-table predictions from the \wdata~\cite{pasupat} training set and \sdata~\cite{tadre}. Calibration excludes evaluation question--table pairs and is performed offline, with the ranges fixed during evaluation.
}

\paragraph{Table Size.}
\info{R2-Q1/5}
\fix{
Consistent with Section~\ref{sec:challenge-analysis}, we measure table size as the token length of the serialized table.
For table $T$ and model $M$, we define
\vspace{-0.6cm}
\begin{center}
\small
\begin{equation}
size_M(T)=\operatorname{Tok}_M\!\left(S(T)\right),
\label{eq:table-size}
\end{equation}
\end{center}
where $S(T)$ is the serialized table and $\operatorname{Tok}_M$ is the tokenizer of model $M$.
This isolates the table-size variation while keeping the prompt template fixed.
}

\paragraph{Reference Accuracy.}
For each model, we use tables with $size_M(T)\leq2{,}048$ tokens as a common small-table reference range, matching the first size range in our empirical analysis.
Let $\widehat{A}^{\mathrm{short}}_M$ denote the model's full-table accuracy within this range.
We allow an accuracy drop of $\epsilon=0.10$ and define
$\tau_M=\widehat{A}^{\mathrm{short}}_M-\epsilon$.
This target specifies the minimum accuracy that larger table-size ranges must reliably maintain.
The same reference range and tolerance are used across models, operation groups. The calibrated limits are reused during evaluation.

\paragraph{Range Calibration.}
\info{R1-Q1/3/5, R2-Q1/5, R3-Q3}
\fix{
For each model $M$ and operation group
$z\in\{\textsc{Light},\textsc{Heavy}\}$,
we independently examine whether full-table performance remains above the reliability target $\tau_M$ as table size increases. We collect the held-out question--table pairs assigned to group $z$ and measure each table by its serialized token size $size_M(T)$. These pairs are ordered by $size_M(T)$ and partitioned into consecutive
$256$-token table-size bins. Each bin therefore contains the question--table pairs whose tables fall within the same token-size interval. Adjacent bins are merged until each resulting calibration bin contains at least
$n_{\min}=50$ examples. For a fixed model--group pair $(M,z)$, let $b$ index a resulting calibration bin.
Let $n_b$ denote the number of examples in bin $b$, and $k_b$ the number answered correctly.
For $k_b>0$, we compute the one-sided Clopper--Pearson lower confidence bound as
\vspace{-0.5cm}
\begin{center}
\small
\begin{equation}
L_{M,z,b}
=
Q_{\mathrm{Beta}}
\left(
\alpha
\mid
k_b,\,
n_b-k_b+1
\right).
\label{eq:lcb}
\end{equation}
\end{center}
When $k_b=0$, we set $L_{M,z,b}=0$.
Here, $Q_{\mathrm{Beta}}$ denotes the Beta-distribution quantile function, whose shape parameters $k_b$ and $n_b-k_b+1$ are determined by the observed outcomes rather than tuned.
We use $\alpha=0.05$, giving a one-sided $95\%$ lower confidence bound.

For each $(M,z)$ pair, starting from the smallest tables, we retain consecutive calibration bins satisfying
$L_{M,z,b}\geq\tau_M$.
The reliable range ends before the first bin whose lower bound falls below $\tau_M$.
The upper endpoint of the retained bins defines $B_M^z$, the upper limit of the reliable table-size range for model $M$ and operation group $z$.
Thus, each base model obtains separate boundaries for the
\textsc{Light} and \textsc{Heavy} groups.

Using the lower confidence bound rather than observed accuracy alone reduces the risk that a small or noisy bin produces an overly optimistic reliable range. All defaults are fixed globally.
The calibrated reliable ranges and routing distributions are reported in
Appendix~\ref{app:routing-calibration}.
}

\paragraph{Route Selection.}
\info{R1-Q1/2, R2-Q1/4/5, R3-Q1}
\fix{
Let $\ell_M(q,T)$ denote the token length of the complete
\readroute{} prompt, including the instruction, question, evidence, serialized table, and output format.
Let $W(M)$ denote the usable context-window budget of model $M$, and $r^*(q,T,M)$ the selected route.

We set $r^*(q,T,M)=\execroute{}$
when $\ell_M(q,T)>W(M)$.
For inputs that fit within the context window, the table size $size_M(T)$ and calibrated range limit $B_M^{z(q,T)}$ determine the route:
\vspace{-0.5cm}
\begin{center}
\small
\begin{equation}
r^*(q,T,M)=
\begin{cases}
\readroute{},
& size_M(T)\leq B_M^{z(q,T)},\\
\reduceroute{},
& size_M(T)>B_M^{z(q,T)}.
\end{cases}
\label{eq:routing}
\end{equation}
\end{center}

Thus, the complete prompt length identifies over-window inputs, while
the calibrated table-size range determines the
\readroute{}--\reduceroute{} decision.
Validated evidence supports the selected route but does not modify the range. After evidence generation, routing requires only operation grouping, token counting, and comparison with the stored range. It requires no additional routing LLM call and uses only one inference route.
}

\subsection{Route Inference}
\label{sec:route-inference}

After routing, \method{} provides the validated evidence and, when needed, the required operations to the selected inference route. Each instance is answered using only that route.

\textbf{\readroute{}}
answers from the full serialized table. Validated evidence is included in the prompt to highlight relevant columns, values, and conditions without removing other table content.

\textbf{\reduceroute{}}
uses validated evidence and required operations to generate a sequence of projection, filtering, neighbor retrieval, sorting, and top-$k$ operations, with aggregation or computation when needed.
After validating their schema and arguments, it applies the sequence to construct a question-focused sub-table for answer generation.

\textbf{\execroute{}}
 uses validated evidence to guide schema linking, value matching, condition construction, and program generation.
The program is applied to the table, and the final answer is obtained from the program output.

\paragraph{Post-Routing Support.}
\fix{
Computational support is generated only after route selection.
When grounded aggregation or arithmetic can be executed deterministically, results such as counts, sums, averages, differences, or ratios are added as optional context for final answer generation.
This support is separate from the operations used by \reduceroute{} to construct a reduced table and does not affect evidence generation, operation grouping, calibration, or routing.
If unavailable or invalid, inference proceeds without it.

Route-specific prompts and fallback rules are provided in Appendix~\ref{app:route-interfaces}.
} \info{R2-Q3}

\begin{table*}[t]
\centering
\scriptsize
\setlength{\tabcolsep}{2.2pt}
\renewcommand{\arraystretch}{1.04}

\begin{minipage}[t]{0.32\textwidth}
\centering
\textbf{WTQ (Mostly Short)}\\[0.18em]
\resizebox{0.96\linewidth}{!}{
\begin{tabular}{lcccc}
\toprule
\textbf{Method} & \textbf{Qwen} & \textbf{LLaMA} & \textbf{GPT} & \textbf{Avg.} \\
\midrule
End-to-End QA   & 42.6 & 51.1 & 56.3 & 50.0 \\
Python QA       & 41.9 & 53.5 & 51.2 & 48.9 \\
GraphOTTER      & 55.3 & 55.1 & 50.3 & 53.6 \\
Binder          & 58.5 & 57.0 & 61.4 & 59.0 \\
Dater           & 60.2 & 63.8 & 59.8 & 61.3 \\
RowColRetrieval      & 53.4 & /    & 55.2 & 54.3 \\
TableRAG        & 59.7 & /    & 55.2 & 57.5 \\
TALON           & 69.6 & 64.6 & 71.0 & 68.4 \\
Table-Critic    & 77.2 & 70.1 & 73.9 & 73.7 \\
H-STAR          & / & 75.8 & 74.9 & 75.4 \\ 
TUNES           & / & 76.8 & \second{75.4} & 76.1 \\ 
RoT             & \second{77.3} & \second{78.7} & 74.2 & \second{76.7} \\
\rowcolor{green!10}
\method\ (ours) & \best{79.1} & \best{79.7} & \best{77.5} & \best{78.8} \\
\rowcolor{green!10}
$\Delta$        & +1.8 & +1.0 & +2.1 & +2.1 \\
\bottomrule
\end{tabular}
}
\end{minipage}
\hfill
\begin{minipage}[t]{0.32\textwidth}
\centering
\textbf{WTQ-L (Mostly Long)}\\[0.18em]
\resizebox{0.96\linewidth}{!}{
\begin{tabular}{lcccc}
\toprule
\textbf{Method} & \textbf{Qwen} & \textbf{LLaMA} & \textbf{GPT} & \textbf{Avg.} \\
\midrule
End-to-End QA   & 34.0 & 40.2 & 55.5 & 43.2 \\
Python QA       & 36.3 & /    & 53.8 & 45.1 \\
Binder          & 60.3 & /    & 61.0 & 60.7 \\
Dater           & 51.2 & /    & 54.7 & 53.0 \\
RowColRetrieval      & 49.9 & /    & 47.2 & 48.6 \\
TableRAG        & 59.6 & /    & 52.1 & 55.9 \\
RoT             & 35.0 & 41.7 & /    & 38.4 \\
TALON           & \second{66.6} & \second{59.5} & \second{67.0} & \second{64.4} \\
\rowcolor{green!10}
\method\ (ours) & \best{68.2} & \best{67.0} & \best{67.6} & \best{67.6} \\
\rowcolor{green!10}
$\Delta$        & +1.6 & +7.5 & +0.6 & +3.2 \\
\bottomrule
\end{tabular}
}
\end{minipage}
\hfill
\begin{minipage}[t]{0.32\textwidth}
\vspace{1.55em}
\centering
\textbf{BirdQA (Mostly Massive)}\\[0.18em]
\resizebox{0.96\linewidth}{!}{
\begin{tabular}{lcccc}
\toprule
\textbf{Method} & \textbf{Qwen} & \textbf{LLaMA} & \textbf{GPT} & \textbf{Avg.} \\
\midrule
Binder          & 59.2 & /    & 60.8 & 60.0 \\
RowColRetrieval     & 42.7 & /    & 48.4 & 45.6 \\
TableRAG        & 53.5 & /    & 54.7 & 54.1 \\
DeepEye-SQL     & 69.1 & \second{72.3} & 69.7 & \second{70.4}\\
TALON           & \second{70.0} & 66.9 & \second{72.0} & 69.6 \\
\rowcolor{green!10}
\method\ (ours) & \best{73.6} & \best{78.7} & \best{74.8} & \best{75.7} \\
\rowcolor{green!10}
$\Delta$        & +3.6 & +6.4 & +2.8 & +5.3 \\
\bottomrule
\end{tabular}
}

\vspace{0.45em}
\footnotesize
Only applicable baselines are shown.
\end{minipage}

\vspace{-0.2em}
\caption{
Main results on \wdata, \ldata, and \bdata. We report exact-match accuracy (\%) under Qwen2.5-72B, LLaMA3.3-70B, and GPT-4o-mini. 
\textbf{Bold} indicates the best and \underline{underlining} indicates the second-best result in each column. 
$\Delta$ denotes the performance gain of \method{} over the best baseline. We report results under the matched setting whenever implementation is available; unavailable or inapplicable entries are marked by ``/''.
}
\label{tab:main}
\end{table*}

\section{Experiments}
\label{sec:experiments}

\info{R1-Q3/5/6, R2-Q2/5/6, R3-Q2/3/4}
\fix{
We evaluate \method{} through three questions.
\textbf{EQ1: Across-scale effectiveness.}
Does \method{} improve TableQA performance across short, long, and massive tables?
\textbf{EQ2: Failure mitigation.}
Does \method{} reduce the grounding, localization, computation, and program errors identified in Section~\ref{sec:challenge-analysis}?
\textbf{EQ3: Generalization and efficiency.}
Does \method{} generalize across datasets and diverse base LLMs while maintaining efficient single-route inference?
}

\subsection{Experimental Setup}

\paragraph{Base Models.}
Our main experiments use three base LLMs from different model families:
Qwen2.5-72B~\citep{qwen}, LLaMA3.3-70B~\citep{llama}, and
GPT-4o-mini~\citep{gpt}.
They cover open- and closed-source settings and different model scales. \info{R3-Q4}
\fix{To broaden the backbone coverage, we also evaluate GPT-5-mini~\cite{gpt5}}. Detailed model and decoding settings are provided in Appendix~\ref{app:exp-settings}.

The route implementations follow Section~\ref{sec:route-inference}, with \execroute{} instantiated using DeepEye-SQL~\citep{li2025deepeye}.

\paragraph{Benchmarks.}
Our primary evaluation for \method{} uses three benchmarks covering different table scales and reasoning settings. \infon{AC-Q2, R1-W2, R3-W1}\fixn{\textbf{WikiTableQuestions} (\wdata)~\citep{pasupat} contains mostly \underline{short tables} with varying lengths. \textbf{WTQ-L}~\citep{talon} derived from \wdata, contains \underline{longer tables} with substantial row-count variation. \textbf{BirdQA}~\citep{talon}, derived from BIRD-SQL~\citep{bird}, is dominated by \underline{massive tables}, with over-context instances requiring program execution.} We further use \textbf{TableBench}~\citep{tabbench} as a
\underline{composite} benchmark for supplementary generalization evaluation, covering fact checking, numerical reasoning, and data analysis. The dataset statistics are shown in Appendix~\ref{dataset}.

\paragraph{Baselines.} We compare our \method{} with four groups of baselines. \textit{Direct answering} includes End-to-End QA with full-table prompting and RoT~\citep{rot} with row-wise traversal.
\textit{Reduction-based methods} include Binder~\citep{binder} with SQL-based decomposition,
Dater~\citep{dater} with evidence and question decomposition, and RowColRetrieval~\citep{talon} with row--column retrieval. \textit{Execution-based methods} include Python QA with code generation, TableRAG~\citep{tabrag} with program-guided retrieval, DeepEye-SQL~\citep{li2025deepeye} with staged Text-to-SQL generation and revision, and TALON~\citep{talon} with multi-agent tool use. \info{R1-Q6, R3-Q2}\fix{\textit{Hybrid methods} include GraphOTTER~\citep{graphotter} with graph reasoning,
Table-Critic~\citep{tablecritic} with critic-based refinement, TUNES~\citep{TUNES} with entity-oriented search, H-STAR~\citep{hstar} with hybrid SQL--text reasoning. All matched baselines use the same models and metrics as \method{}.}

\paragraph{Evaluation Metrics.}
For \wdata, \ldata, and \bdata, we report exact-match (EM) accuracy using the official WikiTableQuestions evaluator~\citep{pasupat}. For the supplementary TableBench evaluation in Appendix~\ref{app:tablebench},
we report ROUGE-L~\citep{rouge}, following prior work~\citep{tabbench,rot}.


\subsection{Main Results (EQ1).}
\info{R1-Q6, R3-Q2}
\fix{}
Table~\ref{tab:main} reports results on \wdata, \ldata, and \bdata.
\method{} achieves the best accuracy across all benchmark--backbone combinations. Averaged over the three base LLMs, it outperforms the strongest baseline by
2.1 points on \wdata{} (RoT), 3.2 points on \ldata{} (TALON), and 5.3 points on \bdata{} (DeepEye-SQL). \infon{AC-Q1, AC-Q2, R1-W1, R1-W2, R1-W4, R2-C2}\fixn{Gains on \ldata{} reflect both routing and evidence conditioning, while \bdata{} improvements primarily reflect evidence-conditioned inference advantage. Against the strongest matched baselines with at least two same backbones, paired McNemar tests show significant improvements on all three benchmarks ($p$ < 0.001).
}

Table~\ref{tab:wtql-row-bucket} in Appendix~\ref{app:row_count_analysis} shows that \method{} achieves the highest accuracy across all four model and row-count settings on \ldata{}, outperforming the strongest baseline by $0.5$ to $2.5$ points.

\begin{table}[t]
\centering
\resizebox{\linewidth}{!}{%
\scriptsize
\setlength{\tabcolsep}{5pt}
\begin{tabular*}{\columnwidth}{@{\extracolsep{\fill}}lrrc>{\columncolor{green!10}}c@{}}
\toprule
Computation Type & \#Q & CoT & \method{} & $\Delta$ \\
\midrule
Count & 1704 & 73.7 & 74.8 & +1.1 \\
Diff. & 103 & 65.0 & 68.9 & +3.9 \\
Sum & 105 & 78.1 & 80.0 & +1.9 \\
Avg. & 23 & 56.5 & 65.2 & +8.7 \\
Extrema & 85 & 72.9 & 76.5 & +3.6 \\
Ratio & 11 & 63.6 & 72.7 & +9.1 \\
\bottomrule
\end{tabular*}
}
\caption{Averaged accuracy by computational question types over base LLMs on \wdata. $\Delta$ denotes absolute gain over CoT. \# Q means number of questions.}
\label{tab:computation-intent}
\end{table}

\subsection{Failure Mitigation Analysis (EQ2)}
\label{sec:failure-mitigation}

\paragraph{Computational questions.}
RQ1 identifies computation errors as a frequent failure mode on short and long tables. Table~\ref{tab:computation-intent} compares \method{} with CoT~\citep{COT} across computational question types. \method{} improves on all listed types, with larger gains on ratio (+9.1), average (+8.7), and difference (+3.9). The smaller gain on counting (+1.1) suggests that precise evidence localization and filtering remain challenging.
Overall, \method{} better handles questions requiring derived computation.

\paragraph{Semantic mismatch.}
We manually analyze semantic mismatch, the largest error category in RQ1. We randomly sample $200$ \wdata{} errors and categorize all errors on \ldata{} and \bdata{}. Semantic mismatch accounts for 31.5\%, 20.2\%, and 34.5\% of the analyzed errors, respectively, compared with 44.0\%, 31.0\%, and 52.0\% in Table~\ref{tab:error-distribution}. Combined with higher accuracy, these lower proportions suggest that \method{} reduces semantic-mismatch errors across table scales.

\subsection{Generalization and Efficiency (EQ3).}
\label{sec:generalization-efficiency}
\paragraph{Generalization.}

Using GPT-5-mini, \method{} outperforms the strongest baseline by $0.8$, $1.7$, and $3.5$ points on \wdata{}, \ldata{}, and \bdata{}, respectively, demonstrating generalization to an additional base LLM. On TableBench, with RoT used as the \readroute{} backbone, \method{} improves ROUGE-L from $44.8$ to $48.7$ under LLaMA3.3-70B. \infon{AC-Q1, AC-Q2, R1-W2, R2-W6} \fixn{These results suggest that the combined evidence-and-routing framework generalizes across backbones, table scales, and table-reasoning tasks.} GPT-5-mini and TableBench results are provided in Appendices~\ref{app:gpt5mini} and~\ref{app:tablebench}.

\begin{figure}[t]
\centering
\includegraphics[width=1\columnwidth]{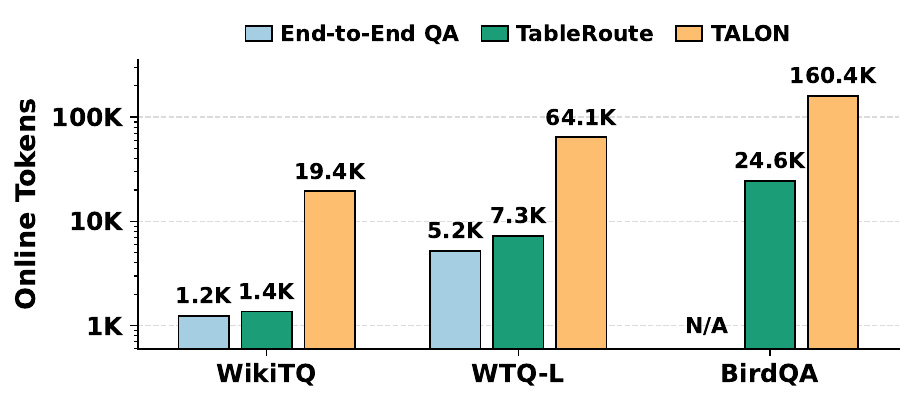}
\caption{
Average total online LLM token consumption per question on a logarithmic scale. Full-table End-to-End QA is not reported for \bdata{} because most inputs exceed the model context window.
} 
\label{fig:token-consumption}
\end{figure}

\paragraph{Token Efficiency.}
Figure~\ref{fig:token-consumption} reports average online LLM token consumption per question. For \method{}, the total includes evidence generation, route-specific planning or program generation, and answer generation. Evidence is generated once per question--table pair, while routing is deterministic and uses no additional LLM call.

\infon{R1-W5, R2-W4, R2-C1, R2-C3}\fixn{Offline calibration is excluded from these totals. The six model–operation limits require 1,952 held-out pairs, 10.17M input tokens, and 79 minutes. Additional latency and evidence-extraction cost analyses are provided in Appendix~\ref{app:additional_efficiency}.}

Including all online stages, \method{} consumes $1.4$K, $7.3$K, and $24.6$K LLM tokens per question on \wdata{}, \ldata{}, and \bdata{}, respectively, compared with $19.4$K, $64.1$K, and $160.4$K for TALON. On \wdata{}, its cost is also close to End-to-End QA (1.2K). Thus, \method{} substantially reduces online inference cost relative to TALON across all three datasets.



\paragraph{Ablation Study.}
\info{R1-Q3, R2-Q3/5, R3-Q3}
\fix{
Table~\ref{tab:ablation} shows that validated evidence is especially important on \ldata{} and \bdata{}, where its removal reduces accuracy by $3.8$ and $4.9$ points.
Computational support contributes $2.0$ and $2.1$ points on \wdata{} and \ldata{}, respectively. \infon{AC-Q1, AC-Q3, R1-W1, R2-W2, R2-W3}\fixn{Fixed-route ablations demonstrate route complementarity; controlled router comparisons in Appendix~\ref{app:router_comparison} isolate the benefit of capability-aware selection. On \ldata{}, per-model calibration raises accuracy from $66.1$ to $66.9$, while \textsc{Light}/\textsc{Heavy} ranges improve it to $67.6$. Analyses of semantic routing, per-operation calibration, and calibration transfer are provided in Appendix~\ref{app:calibration_comparison}.}}

\begin{table}[t]
\centering
\small
\resizebox{\columnwidth}{!}{
\begin{tabular}{lccc}
\toprule
\multicolumn{4}{l}{\textit{(a) Components and fixed routes}} \\
Setting & \wdata{} & \ldata{} & \bdata{} \\
\midrule
\method{}
& \best{78.8} & \best{67.6} & \best{75.7} \\
w/o Validated Evidence
& 76.3 {\scriptsize($\downarrow$2.5)}
& 63.8 {\scriptsize($\downarrow$3.8)}
& 70.8 {\scriptsize($\downarrow$4.9)} \\
w/o Computational Support
& 76.8 {\scriptsize($\downarrow$2.0)}
& 65.5 {\scriptsize($\downarrow$2.1)}
& -- \\
\readroute{}-only
& 75.2 {\scriptsize($\downarrow$3.6)}
& 41.0 {\scriptsize($\downarrow$26.6)}
& -- \\
\reduceroute{}-only
& 73.8 {\scriptsize($\downarrow$5.0)}
& 61.3 {\scriptsize($\downarrow$6.3)}
& -- \\
\execroute{}-only
& 64.6 {\scriptsize($\downarrow$14.2)}
& 59.5 {\scriptsize($\downarrow$8.1)}
& 70.4 {\scriptsize($\downarrow$5.3)} \\
\midrule
\multicolumn{4}{l}{
\textit{(b) Reliable-range calibration on \ldata{}}
} \\
Model & Global & Per-Model & Per-Model+Group \\
\midrule
GPT
& 66.4 {\scriptsize($\downarrow$1.2)}
& 67.0 {\scriptsize($\downarrow$0.6)}
& \textbf{67.6} \\
LLaMA
& 65.5 {\scriptsize($\downarrow$1.5)}
& 66.4 {\scriptsize($\downarrow$0.6)}
& \textbf{67.0} \\
Qwen
& 66.4 {\scriptsize($\downarrow$1.8)}
& 67.3 {\scriptsize($\downarrow$0.9)}
& \textbf{68.2} \\
\midrule
Avg.
& 66.1 {\scriptsize($\downarrow$1.5)}
& 66.9 {\scriptsize($\downarrow$0.7)}
& \textbf{67.6} \\
\bottomrule
\end{tabular}
}
\caption{
Component, fixed-route, and routing-calibration ablations in EM accuracy (\%). Parentheses denote drops from \method{} in (a) and Per-Model+Group in (b). Route-only rows assign applicable instances to the route. Global uses one range across models and groups, Per-Model one per base LLM, and Per-Model+Group separate \textsc{Light} and \textsc{Heavy} ranges per base LLM.
``--'' denotes an inapplicable setting.
}
\label{tab:ablation}
\end{table}

\section{Conclusion}
\label{sec:conclusion}
\infon{AC-Q1, AC-Q2, R1-W1, R1-W2} \fixn{
We propose \method{}, a scalable TableQA framework combining capability-aware routing with evidence-conditioned inference. Given a question, table, and base LLM, it generates validated evidence and required operations, then selects direct answering, table reduction, or program execution using table size and reliability ranges calibrated for each model and operation group. Experiments across datasets, table scales, and base LLMs demonstrate complementary benefits from routing and evidence conditioning, improving accuracy while reducing online token consumption.}

\section*{Limitations}

\method{} requires a one-time calibration to establish reliable ranges for each base LLM.
Its \textsc{Light}/\textsc{Heavy} grouping compresses fine-grained operations into two routing categories and therefore omits distinctions among some operation patterns. We assume  textual or programmatic table access, leaving visually structured tables, noisy spreadsheets, and multi-table documents
with implicit layout relations for future work. 

\section*{Ethics Statement}
All datasets and models used in this paper are publicly available, and their usage strictly adheres to the corresponding licenses and terms of use.

\bibliography{references_arr}

\clearpage
\newpage
\appendix

\section{Comparison with (Multi-)Strategy TableQA Systems}
\label{app:method_comparison}

Table~\ref{tab:method_comparison} summarizes the differences between \textsc{TableRoute} and representative multi-strategy or long-table TableQA systems. The goal of this comparison is not to claim that one framework strictly subsumes another, but to clarify the design choices that distinguish \textsc{TableRoute}. In particular, \textsc{TableRoute} performs pre-execution single-route selection, uses a calibrated model-specific routing boundary, and reuses the same validated evidence across direct answering, table reduction, and program execution.

\begin{table*}[ht]
\centering
\small
\resizebox{\textwidth}{!}{
\setlength{\tabcolsep}{4pt}
\begin{tabular}{lccccccc}
\toprule
Method 
& Direct QA 
& Table Reduction 
& Program Execution 
& Pre-exec. Routing 
& Single-route Inference 
& Validated Evidence 
& Model-specific Boundary \\
\midrule
Binder 
& \xmark 
& \cmark 
& \cmark 
& \xmark 
& \cmark 
& \xmark 
& \xmark \\
Dater 
& \xmark 
& \cmark 
& \xmark 
& \xmark 
& \cmark 
& \xmark 
& \xmark \\
Chain-of-Table 
& \xmark 
& \cmark 
& \xmark 
& \xmark 
& \cmark 
& \xmark 
& \xmark \\
SynTQA 
& \cmark 
& \xmark 
& \cmark 
& \xmark 
& \xmark 
& \xmark 
& \xmark \\
TableRAG 
& \xmark 
& \cmark 
& \cmark 
& \xmark 
& \xmark 
& \xmark 
& \xmark \\
TALON 
& \xmark 
& \cmark 
& \cmark 
& \xmark 
& \xmark 
& \xmark 
& \xmark \\
TableMaster 
& \cmark 
& \cmark 
& \cmark 
& \cmark 
& \xmark 
& \xmark 
& \xmark \\
\textsc{\method{}} 
& \cmark 
& \cmark 
& \cmark 
& \cmark 
& \cmark 
& \cmark 
& \cmark \\
\bottomrule
\end{tabular}
}
\caption{
Comparison between \textsc{TableRoute} and representative multi-strategy or long-table TableQA systems.
Direct QA denotes full-table answer generation.
Table reduction denotes explicit construction of a smaller sub-table.
Program execution denotes SQL-, Python-, or tool-based answer execution.
Pre-execution routing means that the route is selected before answer generation.
Single-route inference means that only one selected route is executed for each question.
Validated evidence denotes typed and executable question--table grounding that is checked before being used by downstream routes.
Model-specific boundary denotes a calibrated routing threshold that depends on the base model capability and required operations.
}
\label{tab:method_comparison}
\end{table*}

\section{Details for Empirical Analysis}
\label{app:challenge-analysis}

This appendix provides details for the diagnostic analysis in Section~\ref{sec:challenge-analysis}. 
The goal of this analysis is to motivate the design of \method{}, rather than to serve as the final benchmark evaluation. 
Final benchmark results are reported in Section~\ref{sec:experiments}.

\subsection{Failure Categorization Details}
\label{app:error-categorization}

For RQ1, we analyze 200 randomly sampled incorrect cases for each table-scale regime. 
Across all regimes, we include TALON~\citep{talon} as a representative long-table exploration baseline. 
We additionally include CoT-style full-table answering~\citep{COT} for short tables, RoT~\citep{rot} for long tables, and DeepEye-SQL~\citep{li2025deepeye} for massive tables. 
For each incorrect case, we identify the first responsible failure and categorize it into one of the following groups.

\paragraph{Semantic mismatch.}
This category includes failures where the model or generated program maps the question to the wrong table semantics. 
Typical cases include selecting the wrong column, matching the wrong cell value, confusing semantically similar headers, misinterpreting a constraint, or using an incorrect operation implied by the question.

\paragraph{Computation error.}
This category includes failures where the relevant evidence is correctly identified but the numerical computation is wrong through LLM reasoning. 
Examples include incorrect counting, aggregation, comparison, difference, ratio, percentage, or arithmetic over table values.

\paragraph{Evidence localization error.}
This category includes failures where the model fails to locate the required evidence in the table. 
Examples include overlooking relevant rows, columns, or cells, attending to an irrelevant table region, or failing to retrieve the row needed for a comparison or lookup.

\paragraph{Reduction or execution error.}
This category includes failures introduced after a reduction or execution-style route is chosen. 
Examples include incomplete reduced tables, invalid SQL or Python programs, incorrect filters, wrong aggregation functions, missing ordering constraints, or erroneous program parameters.

\paragraph{Others.}
This category includes failures that do not fit the above groups, such as formatting failures, or cases where multiple errors are inseparable.

\paragraph{Massive-table cases.}
For massive tables, answers are obtained mainly through executable operations rather than direct inspection of the full serialized table. 
Therefore, localization or computation failures caused by generated executable operations are counted as reduction or execution errors. 
This convention explains why computation and evidence-localization entries are marked as ``--'' for massive tables in Table~\ref{tab:error-distribution}.

\paragraph{Table-scale regimes and token bins.}
In Section~\ref{sec:challenge-analysis}, table size is measured by the number of tokens in the serialized table. 
We use three benchmark-scale regimes: \textit{short tables}, which can be serialized and directly read with limited irrelevant context; \textit{long tables}, which still fit within the model context window but make full-table reading less reliable due to additional rows or columns; and \textit{massive tables}, whose serialized form exceeds the model context window and therefore requires execution-based access. 
This benchmark-scale taxonomy is distinct from the calibration token bins used in Figure~\ref{fig:size-effects}(b) and Appendix~\ref{app:length-operation-effects}. 
For the calibration analysis, we group context-fit examples into \textit{Short} ($\leq 2048$ tokens), \textit{Mid} ($2560$--$6144$ tokens), and \textit{Long} ($\geq 7168$ tokens) bins to estimate how full-table answering degrades with table length and question type.

\subsection{Execution Baselines for Massive Tables}
\label{app:massive-execution-baselines}

For massive tables, full-table prompting is infeasible and answers are obtained mainly through executable operations. 
Table~\ref{tab:massive-text2sql} reports representative Text-to-SQL results on \bdata. 
Although execution avoids placing the full table in the prompt, performance remains sensitive to schema grounding, value matching, and operation construction. 
This supports the RQ1 observation that massive-table failures are dominated by semantic mismatch and reduction or execution errors.

\begin{table}[h]
\centering
\small
\setlength{\tabcolsep}{6pt}
\begin{tabular*}{\columnwidth}{@{\extracolsep{\fill}}lr@{}}
\toprule
Method & Acc. \\
\midrule
Bird~\citep{bird} & 55.8 \\
DAIL-SQL~\citep{DAILSQL} & 54.8 \\
TA-SQL~\citep{tqsql} & 56.1 \\
MAC-SQL~\citep{macsql} & 59.6 \\
DeepEye-SQL~\citep{li2025deepeye} & 69.7 \\
\bottomrule
\end{tabular*}
\caption{Representative Text-to-SQL results on \bdata{} for massive-table settings using GPT-4o-mini.}
\label{tab:massive-text2sql}
\end{table}

\section{Evidence Generation Details}
\label{app:evidence-generation}

This appendix describes the implementation of the evidence generation module in Section~\ref{sec:evidence}. 
The goal is to obtain lightweight, verifiable question--table alignments without serializing the full table into the LLM prompt.

\subsection{Compact Table Profile}
\label{app:compact-profile}

For each question-table pair, we construct a \textit{compact table profile} $P(T)$, which is a bounded-size summary of the table used only for evidence extraction. 
It contains four components: 
(1) all column names; 
(2) an inferred type for each column, chosen from \texttt{text}, \texttt{number}, and \texttt{date}; 
(3) up to five head non-null values per column; and 
(4) up to five representative rows.

The value and row budgets are set to five to keep the profile short while still exposing common surface forms to the evidence extractor. 
We use head non-null values because they are deterministic and inexpensive to obtain. 
Representative rows are selected by lexical overlap with the question. 
Specifically, each row is converted into a string by concatenating its cell values, and the row is scored by the number of shared normalized tokens with the question. 
Here, \textit{normalized tokens} are lowercased whitespace tokens after removing simple punctuation. 
The top-scoring rows are selected; if all rows have zero overlap, we use the first five rows as a deterministic fallback.

For large tables, the LLM only observes $P(T)$ rather than the full serialized table. 
The later validation step is applied to the table view available to the pipeline. 
A \textit{table view} refers to the in-memory table or retrieved table slice that the current route can access; for short and long tables this is typically the full table, while for massive tables it may be a database-backed or retrieved view.

\subsection{Column Type Inference}
\label{app:column-type-inference}

Column types are inferred with a lightweight parser. 
For each column, we consider its non-empty cells. 
A column is assigned type \texttt{number} if at least 80\% of its non-empty cells can be parsed as numbers, and type \texttt{date} if at least 80\% can be parsed as dates. 
If neither condition holds, the column is treated as \texttt{text}. 
The 80\% threshold is used to avoid classifying mixed or noisy columns as numeric or date columns due to a few parseable values.

For number parsing, comma separators are removed before numeric conversion. 
For date parsing, invalid values are coerced to missing values. 
The current implementation does not perform full semantic normalization for currencies, percentages, units, or domain-specific date ranges; these cases may be filtered out during validation or handled by downstream route-specific reasoning.

\subsection{Structured Evidence Schema and Prompt}
\label{app:evidence-schema-prompt}

The evidence extractor uses a zero-shot instruction prompt with temperature set to 0 for deterministic structured output. 
The model receives the question and compact table profile, and returns a list of evidence candidates. 
Each candidate contains the fields in Table~\ref{tab:evidence-schema}.

\begin{table}[h]
\centering
\small
\setlength{\tabcolsep}{4pt}
\begin{tabular}{lp{0.58\columnwidth}}
\toprule
Field & Description \\
\midrule
\texttt{semantic\_unit} & Question phrase to be grounded. \\
\texttt{column\_hint} & Natural-language hint for the target column. \\
\texttt{grounding\_type} & Expected value type: \texttt{text}, \texttt{number}, or \texttt{date}. \\
\texttt{operator} & Condition operator. \\
\texttt{value} & Main literal value. \\
\texttt{value2} & Second value for \texttt{between}; empty otherwise. \\
\bottomrule
\end{tabular}
\caption{Evidence candidate schema.}
\label{tab:evidence-schema}
\end{table}

The supported operators are equality, inequality, comparison, containment, and range matching:
\texttt{==}, \texttt{!=}, \texttt{>}, \texttt{>=}, \texttt{<}, \texttt{<=}, \texttt{contains}, and \texttt{between}.

A compact version of the prompt template is shown below. 
Line breaks are added for readability.

{\scriptsize
\begin{quote}
\ttfamily
Given a question and a compact table profile,\\
extract evidence items that link question spans\\
to table columns and values.\\[2pt]
Return only a JSON list. Each item contains:\\
semantic\_unit, column\_hint, grounding\_type,\\
operator, value, and value2.\\[2pt]
grounding\_type: text, number, or date.\\
operator: ==, !=, $>$, $>=$, $<$, $<=$, contains, between.\\[2pt]
Question: \{question\}\\
Table profile: \{table\_profile\}
\end{quote}
}

If JSON parsing fails, we retry once with a shorter prompt that preserves the schema but removes explanatory instructions. 
A single retry is used to recover from formatting errors without introducing substantial extra cost. 
If parsing still fails, the evidence list is set to empty and downstream inference proceeds without evidence. 
The current implementation focuses on typed row-selection evidence; aggregation and arithmetic are handled by route-specific reasoning modules rather than represented as executable evidence items.

\subsection{Column Resolution}
\label{app:column-resolution}

Column resolution maps a generated \texttt{column\_hint} to an actual table column. 
For each candidate, we first remove columns whose inferred type is incompatible with the candidate's \texttt{grounding\_type}. 
We then score each remaining column by lexical similarity between the normalized hint $h$ and normalized column header $c$:
{
\begin{align}
s(h,c) &= \mathrm{SeqMatch}(h,c) \notag \\
&\quad + \lambda_{\mathrm{sub}} \cdot \mathrm{Substr}(h,c).
\end{align}
}
Here, $\mathrm{SeqMatch}(h,c)$ is the SequenceMatcher ratio, and $\mathrm{Substr}(h,c)$ is 1 if either string is a substring of the other and 0 otherwise. 
We set $\lambda_{\mathrm{sub}}=0.3$ to reward exact containment, which often occurs when a generated hint is a shortened or expanded version of a table header.

The best-scoring column is selected if its score is at least 0.35; otherwise, the evidence is marked \textit{unresolved} and dropped. 
The threshold is selected on the calibration split and then fixed for all evaluations. 
A relatively low threshold is used because the evidence is later checked by typed execution, so column resolution is allowed to be recall-oriented while validation controls precision.

\subsection{Typed Validation}
\label{app:typed-validation}

Typed validation checks whether a resolved evidence item can be executed against the resolved column. 
This step is used to filter hallucinated or unsupported evidence before it is passed to downstream routes.

For \texttt{text} columns, we support \texttt{==}, \texttt{!=}, and \texttt{contains} with case-insensitive comparison. 
For \texttt{number} and \texttt{date} columns, we support \texttt{==}, \texttt{!=}, \texttt{>}, \texttt{>=}, \texttt{<}, \texttt{<=}, and \texttt{between}. 
An evidence item is dropped if the column cannot be resolved, the inferred type is incompatible with the evidence type, the evidence value cannot be parsed, the operator is unsupported for the resolved type, or the condition cannot be executed. 
We refer to the last two cases as \textit{not executable}.

An executable condition is not dropped solely because it matches zero rows. 
Instead, it is retained with an \textit{empty-match} flag. 
An empty-match item is used only as a \textit{soft constraint}, meaning that it can inform prompting, decomposition, or program repair, but it is not applied as a mandatory filter that would remove all rows. 
This design avoids over-filtering when the evidence is semantically plausible but fails due to surface-form mismatch or limited normalization.

\subsection{Route Integration and Cost Accounting}
\label{app:evidence-route-cost}

The same retained evidence list is shared across all routes. 
In \textsc{Direct}, evidence is inserted as auxiliary grounding context. 
In \textsc{Reduce}, executable conditions are converted into filtering or decomposition constraints; a \texttt{between} condition is decomposed into lower- and upper-bound constraints. 
In \textsc{Execute}, evidence constrains column selection, value grounding, and program repair. 
If no evidence is retained, the selected route proceeds without evidence.

Evidence generation is performed once per question-table pair. 
All input and output tokens used for evidence generation are included in the reported token cost.

\subsection{Evidence Statistics}
\label{app:evidence-statistics}

Table~\ref{tab:evidence-summary} reports summary statistics for evidence generation. 
\textit{Avg cand.} is the average number of evidence candidates generated per question. 
\textit{Kept} and \textit{Dropped} are rates over generated candidates. 
\textit{Empty-match} is computed among retained evidence items. 

\begin{table}[h]
\centering
\small
\setlength{\tabcolsep}{5pt}
\begin{tabular}{lrrrr}
\toprule
Dataset & Avg cand. & Kept & Dropped & Empty \\
\midrule
WikiTQ & 1.872 & 0.736 & 0.264 & 30.1\% \\
WTQ-L & 1.773 & 0.811 & 0.189 & 15.5\% \\
\bottomrule
\end{tabular}
\caption{Summary statistics for evidence generation.}
\label{tab:evidence-summary}
\end{table}

Table~\ref{tab:evidence-failures} reports validation failure counts. 
\textit{Unres. col.} counts candidates whose column hint cannot be mapped to a table column. 
\textit{Type mis.} counts candidates removed because the evidence type is incompatible with the resolved column type. 
\textit{Parse fail} counts candidates whose value cannot be parsed under the predicted type. 
\textit{Not exec.} counts candidates with unsupported operators or execution failures. 
For WTQ-L, all values are averaged over the three base models.

Table~\ref{tab:evidence-failures} reports validation failure counts and their rates among generated evidence candidates. 
\textit{Unresolved column} means that the column hint cannot be mapped to a table column. 
\textit{Type mismatch} means that the evidence type is incompatible with the resolved column type. 
\textit{Parse failure} means that the value cannot be parsed under the predicted type. 

\begin{table}[h]
\centering
\small
\setlength{\tabcolsep}{5pt}
\begin{tabular}{lcc}
\toprule
Failure type & WikiTQ & WTQ-L \\
\midrule
Unresolved column & 1180 / 8132 & 56 / 596 \\
Type mismatch & 105 / 8132 & 14 / 596 \\
Parse failure & 600 / 8132 & 34 / 596 \\
Not executable & 971 / 8132 & 57 / 596 \\
\bottomrule
\end{tabular}
\caption{Validation failure breakdown among generated evidence candidates.}
\label{tab:evidence-failures}
\end{table}

\begin{figure}[t]
\centering
\begin{promptbox}{Evidence Generation Prompt}
Given a question and a compact table profile, extract:
(1) evidence linking question spans to table columns
    and conditions;
(2) the operations required to answer the question.

Do not answer the question or invent unsupported
columns or values.

Each evidence item contains:
semantic_unit, column_hint, grounding_type,
operator, value, and value2.

grounding_type: text, number, or date
operator: ==, !=, >, >=, <, <=, contains, between

Required operations must be selected from:
Projection, Filter, Neighbor, Sort, TopK,
Aggregate, Compute.

Return only valid JSON:
{
  "evidence": [
    {
      "semantic_unit": "...",
      "column_hint": "...",
      "grounding_type": "...",
      "operator": "...",
      "value": "...",
      "value2": null
    }
  ],
  "required_operations": ["..."]
}

Question:
{question}

Table profile:
{table_profile}
\end{promptbox}
\caption{Prompt for evidence and required-operation generation.}
\label{fig:evidence-generation-prompt}
\end{figure}

\subsection{Evidence Generation}
\info{R1-Q4, R2-Q2/3}
\fix{
\paragraph{Prompt.} Figure~\ref{fig:evidence-generation-prompt} shows the prompt used to generate evidence candidates and required operations.
}

\paragraph{Evidence Example.}
Figure~\ref{fig:evidence-example} illustrates the complete evidence generation process.
The evidence extractor observes only the question and compact table profile,
then produces evidence candidates and required operations. The candidates are subsequently resolved and validated against the table.

\lstset{
  basicstyle=\ttfamily\scriptsize,
  breaklines=true,
  columns=fullflexible,
  keepspaces=true,
  showstringspaces=false
}


\begin{figure*}[t]
\centering

\begin{minipage}[t]{0.55\textwidth}
\vspace{0pt}
\begin{tcblisting}{
  listing only,
  title={Input and Generated Output},
  colback=gray!4,
  colframe=gray!45,
  colbacktitle=gray!25,
  coltitle=black,
  fonttitle=\bfseries,
  boxrule=0.5pt,
  arc=2pt,
  left=1.5mm,
  right=1.5mm,
  top=1mm,
  bottom=1mm,
  before skip=0pt,
  after skip=0pt,
  equal height group=evidence-example,
  valign=top,
  listing options={
    basicstyle=\ttfamily\scriptsize,
    breaklines=true,
    columns=fullflexible,
    keepspaces=true,
    showstringspaces=false
  }
}
Question:
How many Australian teams scored above 10?

Compact table profile:

Columns and head values:
- Team [text]: Melbourne City, Sydney FC,
  Perth Glory, Auckland FC, Wellington Phoenix
- Country [text]: Australia, Australia,
  Australia, New Zealand, New Zealand
- Score [number]: 12, 15, 8, 14, 11

Representative rows:
Melbourne City     | Australia   | 12
Sydney FC          | Australia   | 15
Perth Glory        | Australia   | 8
Auckland FC        | New Zealand | 14
Wellington Phoenix | New Zealand | 11

Generated output:
{
  "evidence": [
    {
      "semantic_unit": "Australian",
      "column_hint": "Country",
      "grounding_type": "text",
      "operator": "==",
      "value": "Australia",
      "value2": null
    },
    {
      "semantic_unit": "above 10",
      "column_hint": "Score",
      "grounding_type": "number",
      "operator": ">",
      "value": "10",
      "value2": null
    }
  ],
  "required_operations": [
    "Filter",
    "Aggregate"
  ]
}
\end{tcblisting}
\end{minipage}
\hfill
\begin{minipage}[t]{0.43\textwidth}
\vspace{0pt}
\begin{tcblisting}{
  listing only,
  title={Validation and Retained Evidence},
  colback=gray!4,
  colframe=gray!45,
  colbacktitle=gray!25,
  coltitle=black,
  fonttitle=\bfseries,
  boxrule=0.5pt,
  arc=2pt,
  left=1.5mm,
  right=1.5mm,
  top=1mm,
  bottom=1mm,
  before skip=0pt,
  after skip=0pt,
  equal height group=evidence-example,
  valign=top,
  listing options={
    basicstyle=\ttfamily\scriptsize,
    breaklines=true,
    columns=fullflexible,
    keepspaces=true,
    showstringspaces=false
  }
}
Column resolution and typed validation:

1. "Country" -> Country
   Type: text
   Condition: == Australia
   Executable: yes
   Matches: 3 rows
   Status: retained

2. "Score" -> Score
   Type: number
   Condition: > 10
   Executable: yes
   Matches: 4 rows
   Status: retained

Validated evidence:

E(q,T) = {
  ("Australian", Country, == Australia),
  ("above 10", Score, > 10)
}

Required operations:

O(q,T) = {Filter, Aggregate}

Operation group:

z(q,T) = Heavy

The evidence stage does not produce the final
answer. It returns validated grounding and
required operations for routing and inference.
\end{tcblisting}
\end{minipage}

\caption{
An example of evidence and required-operation generation.
The extractor produces structured candidates from the compact table profile.
Validation resolves column hints and checks type compatibility and
executability before retaining the evidence.
}
\label{fig:evidence-example}
\end{figure*}

\section{Routing Calibration Details}
\label{app:routing-calibration}

\fix{
This appendix provides implementation details for the capability-aware
routing mechanism in Section~\ref{sec:routing}.
We describe the calibration data, the effects of table size and required
operations, the construction of operation groups, the reliable-range
calibration procedure, and the routing ablations.
}\info{R1-Q1/2/3/5, R2-Q1/4/5, R3-Q1/3}

\subsection{Calibration Data and Table Size}
\label{app:base-boundary}

We calibrate the routing ranges using held-out examples from the \wdata~\cite{pasupat}
training set and SLQA~\cite{tadre}.
\wdata~mainly provides short and medium tables, while SLQA supplies additional large-table examples.
SLQA is used only for calibration and is excluded from all reported test sets.

Following Section~\ref{sec:routing}, we measure table size by the serialized-table token count $size_M(T)$ under model $M$.
We use $size_M(T)\leq S_0$ as the small-table reference range and define
$\tau_M=\widehat{A}^{\mathrm{short}}_M-\epsilon$.
We set $S_0=2{,}048$ tokens, matching the first size range in our empirical analysis, and $\epsilon=0.10$.
These settings are shared across models, operation groups, and evaluation datasets.

\subsection{Empirical Motivation: Length and Operation Effects}
\label{app:length-operation-effects}

For descriptive analysis, we divide context-fit examples into short
($\leq2{,}048$ table tokens), medium
($2{,}049$--$6{,}144$), and long ($>6{,}144$) regions.
These regions are used only to summarize performance trends and are distinct
from the finer calibration bins introduced below.

Across the three base models, direct-route accuracy decreases from
$76.33\%$ in the short region to $63.90\%$ in the long region.
Table~\ref{tab:model-size-effect} reports the trend by model.

\begin{table}[h]
\centering
\small
\setlength{\tabcolsep}{5pt}
\begin{tabular}{lccc}
\toprule
Model & Short & Medium & Long \\
\midrule
Qwen  & 0.750 & 0.656 & 0.617 \\
GPT   & 0.796 & 0.722 & 0.642 \\
LLaMA & 0.740 & 0.644 & 0.658 \\
Avg.  & 0.763 & 0.674 & 0.639 \\
\bottomrule
\end{tabular}
\caption{
Direct-route accuracy across table-size regions.
GPT, LLaMA, and Qwen denote GPT-4o-mini, LLaMA3.3-70B, and
Qwen2.5-72B, respectively.
}
\label{tab:model-size-effect}
\end{table}

Required operations also exhibit different full-table performance. For this descriptive analysis, each example is assigned its most demanding identified operation type.
Table~\ref{tab:operation-main-effect} reports the pooled accuracy and number of examples for each type.

\begin{table}[h]
\centering
\small
\setlength{\tabcolsep}{6pt}
\begin{tabular}{lrr}
\toprule
Operation & Acc. & $n$ \\
\midrule
Projection & 0.736 & 1,214 \\
Sort       & 0.729 &   380 \\
Top-k      & 0.726 &   357 \\
Neighbor   & 0.720 &   214 \\
Filter     & 0.686 &   948 \\
Compute    & 0.686 &   102 \\
Aggregate  & 0.614 &   817 \\
\bottomrule
\end{tabular}
\caption{Average direct-route accuracy by required operation type.}
\label{tab:operation-main-effect}
\end{table}

Table~\ref{tab:operation-length-effect} further reports the interaction
between operation type and table size.
Aggregation and computation show the largest declines, while retrieval-oriented
operations exhibit smaller reductions.

\begin{table}[h]
\centering
\small
\setlength{\tabcolsep}{4.5pt}
\begin{tabular}{lrrr}
\toprule
Operation & Short & Long & $\Delta$ \\
\midrule
Projection & 0.785 & 0.687 & -0.098 \\
Filter     & 0.779 & 0.610 & -0.169 \\
Neighbor   & 0.763 & 0.691 & -0.072 \\
Sort       & 0.781 & 0.648 & -0.133 \\
Top-k      & 0.793 & 0.612 & -0.181 \\
Aggregate  & 0.747 & 0.449 & -0.298 \\
Compute    & 0.889 & 0.526 & -0.363 \\
\bottomrule
\end{tabular}
\caption{
Operation-by-table-size effect.
$\Delta$ denotes long minus short accuracy.
}
\label{tab:operation-length-effect}
\end{table}

\subsection{Operation Groups}
\label{app:column-signal-ratio}
\label{app:question-complexity}

The operation groups follow the definition in
Section~\ref{sec:evidence} and are summarized in
Table~\ref{tab:complexity-prior}.

\begin{table}[h]
\centering
\small
\renewcommand{\arraystretch}{0.95}
\setlength{\tabcolsep}{3pt}
\begin{tabular}{@{}lp{0.66\columnwidth}@{}}
\toprule
Group & Operation types \\
\midrule
\textsc{Light}
& \textsc{Projection}, \textsc{Filter}, \textsc{Neighbor} \\
\textsc{Heavy}
& \textsc{Sort}, \textsc{TopK}, \textsc{Aggregate},
  \textsc{Compute} \\
\bottomrule
\end{tabular}
\caption{Operation groups used for reliable-range.}
\label{tab:complexity-prior}
\end{table}

For each base model, held-out question--table pairs are divided into
\textsc{Light} and \textsc{Heavy} subsets before calibration.
The subsets are calibrated independently, yielding
$B_M^{\textsc{Light}}$ and $B_M^{\textsc{Heavy}}$.
The original operation types remain available when explicitly required by
the selected inference route.

\subsection{Reliable Range Calibration}
\label{app:elastic-hparams}

Following Section~\ref{sec:routing}, we calibrate reliable table-size
ranges independently for each base model and operation group.
Within each $(M,z)$ pair, held-out question--table pairs are ordered by
their serialized table size $size_M(T)$ and initially divided into
$256$-token bins.
Adjacent bins are merged until each contains at least
$n_{\min}=50$ examples.
We then apply the lower-confidence-bound criterion in
Equation~\ref{eq:lcb} to obtain the upper endpoint $B_M^z$ of each
reliable range.

All calibration settings are shared across models and operation groups,
as summarized in Table~\ref{tab:elastic-hparams}.

\begin{table}[h]
\centering
\small
\renewcommand{\arraystretch}{0.95}
\setlength{\tabcolsep}{4pt}
\begin{tabular}{lcc}
\toprule
Setting & Symbol & Default \\
\midrule
Reference table size & $S_0$      & 2,048 \\
Accuracy tolerance   & $\epsilon$ & 0.10 \\
Initial bin width    & --         & 256 \\
Minimum bin size     & $n_{\min}$ & 50 \\
One-sided tail       & $\alpha$   & 0.05 \\
\bottomrule
\end{tabular}
\caption{
Default settings for reliable-range calibration.
}
\label{tab:elastic-hparams}
\end{table}

Table~\ref{tab:calibrated-ranges} reports the six resulting reliable
ranges.
Because table size is measured using each model's tokenizer, the range
limits are interpreted separately for each model.

\begin{table}[h]
\centering
\small
\renewcommand{\arraystretch}{0.95}
\setlength{\tabcolsep}{3pt}
\resizebox{\columnwidth}{!}{
\begin{tabular}{lcccc}
\toprule
Model
& $\widehat{A}^{\mathrm{short}}_M$
& $\tau_M$
& \textsc{Light} range
& \textsc{Heavy} range \\
\midrule
GPT
& 0.796
& 0.696
& $size_M(T)\leq4{,}608$
& $size_M(T)\leq3{,}072$ \\

LLaMA
& 0.740
& 0.640
& $size_M(T)\leq4{,}352$
& $size_M(T)\leq3{,}328$ \\

Qwen
& 0.752
& 0.652
& $size_M(T)\leq4{,}096$
& $size_M(T)\leq2{,}816$ \\
\bottomrule
\end{tabular}
}
\caption{
Calibrated reliable table-size ranges for each base model and operation
group.
Range limits are measured in serialized table tokens.
}
\label{tab:calibrated-ranges}
\end{table}

The shorter \textsc{Heavy} ranges reflect the greater sensitivity of
comparison- and computation-intensive questions to increasing table size.
Calibration is performed once offline.
At inference time, the router only identifies the operation group, counts
table tokens, and compares the resulting size with the stored range limit.
No additional LLM call is required for route scoring.

\subsection{Route Distribution}
\label{app:route-distribution}

Table~\ref{tab:route-dist-fixed-elastic} compares model-only calibration
with the full model- and operation-group-specific calibration on WTQ-L.
The latter routes more context-fit examples to \reduceroute{}, while the
over-window \execroute{} proportion remains unchanged.

\begin{table}[h]
\centering
\small
\setlength{\tabcolsep}{3.5pt}
\begin{tabular}{llrrr}
\toprule
Model & Calibration & Dir. & Red. & Exec. \\
\midrule
GPT
& Model-only  & 48.81 & 42.56 & 8.63 \\
GPT
& Model+Group & 44.35 & 47.02 & 8.63 \\
LLaMA
& Model-only  & 41.97 & 49.40 & 8.63 \\
LLaMA
& Model+Group & 40.48 & 50.89 & 8.63 \\
Qwen
& Model-only  & 46.73 & 44.64 & 8.63 \\
Qwen
& Model+Group & 38.99 & 52.38 & 8.63 \\
\bottomrule
\end{tabular}
\caption{
Route distribution on WTQ-L (\%).
Model-only uses one reliable range per model, while Model+Group uses
separate \textsc{Light} and \textsc{Heavy} ranges.
}
\label{tab:route-dist-fixed-elastic}
\end{table}

\infon{AC-Q1, AC-Q3, R1-W1, R2-W2, R2-W3}\fixn{
\subsection{Controlled Router Comparisons}
\label{app:router_comparison}

To isolate the contribution of capability-aware routing, we compare \method{} against three alternative routing strategies: length-only, LLM-based, and learned routing. All methods use GPT-5-mini with identical evidence conditioning and inference backbones, differing only in route selection.

\begin{table}[t]
\centering
\small
\begin{tabular}{lcc}
\toprule
Routing Strategy & \wdata{} & \ldata{} \\
\midrule
Length-only     & 81.7 & 68.5 \\
LLM-based       & 82.0 & 68.8 \\
Learned         & 82.2 & 69.0 \\
\midrule
\method{}       & \textbf{82.5} & \textbf{69.6} \\
\bottomrule
\end{tabular}
\caption{Controlled router comparison using GPT-5-mini. All strategies share the same evidence conditioning and inference backbones. Results are reported in EM accuracy (\%).}
\label{tab:router_comparison}
\end{table}

As shown in Table~\ref{tab:router_comparison}, \method{} consistently outperforms the alternative routers, improving over length-only routing by 0.8 and 1.1 points on \wdata{} and \ldata{}, respectively. The improvement further increases to 2.7 points on long-table questions requiring HEAVY operations.

To separate routing from evidence conditioning, we additionally compare routing strategies without the shared evidence layer. \method{} still outperforms length-only routing by 0.7 and 0.9 points on \wdata{} and \ldata{}, respectively. These results demonstrate that capability-aware selection provides an independent, albeit modest, contribution beyond evidence conditioning.
}

\infon{}
\subsection{Calibration Transfer and Alternative Routing Strategies}
\label{app:calibration_comparison}

We further evaluate the transferability of calibrated routing boundaries and compare alternative routing strategies. Table~\ref{tab:calibration_comparison} summarizes the results.

\begin{table}[t]
\centering
\small
\begin{tabular}{lcc}
\toprule
Setting & \wdata{} & \ldata{} \\
\midrule
\multicolumn{3}{l}{\textit{Original three backbones (average)}} \\
Model-specific calibration & 78.8 & 67.6 \\
Cross-model calibration    & 78.6 & 67.3 \\
LLM semantic routing       & 79.0 & 68.6 \\
\midrule
\multicolumn{3}{l}{\textit{GPT-5-mini}} \\
Per-operation calibration  & 82.4 & 69.4 \\
\textsc{Light}/\textsc{Heavy} calibration & \textbf{82.5} & \textbf{69.6} \\
Oracle routing             & 84.9 & 76.2 \\
\bottomrule
\end{tabular}
\caption{Calibration transfer and alternative routing strategies on \wdata{} and \ldata{}. Results are reported in EM accuracy (\%). Oracle routing is an analysis-only upper bound.}
\label{tab:calibration_comparison}
\end{table}

\paragraph{Calibration Transfer.}
We examine cross-model transfer by calibrating each of the three original backbones using data from the other two. This setting achieves 78.6\% and 67.3\% accuracy on \wdata{} and \ldata{}, respectively, compared with 78.8\% and 67.6\% under model-specific calibration. The small differences suggest that calibrated boundaries transfer reasonably well across the evaluated model families.

\paragraph{Semantic Routing.}
We compare our lightweight calibrated router with an LLM-based semantic scorer. The semantic scorer achieves 79.0\% and 68.6\% on \wdata{} and \ldata{}, improving over our router by 0.2 and 1.0 points, respectively. However, it requires an additional LLM call for route selection. Our approach therefore offers a favorable accuracy--cost trade-off.

\paragraph{Per-Operation Calibration.}
We further compare the \textsc{Light}/\textsc{Heavy} grouping with finer-grained calibration over seven operation types using GPT-5-mini. Per-operation calibration achieves 82.4\% and 69.4\% on \wdata{} and \ldata{}, slightly below the two-group approach (82.5\% and 69.6\%). This suggests that finer-grained calibration provides no additional benefit under the evaluated setting, potentially due to sparser calibration subsets. An oracle selecting a correct output whenever any candidate route succeeds achieves 84.9\% and 76.2\%, indicating remaining room for improved route selection.

\section{Route-Specific Interface Details}
\label{app:route-interfaces}

We report implementation details for the route-specific interfaces described earlier. 
After routing, \method{} adapts retained evidence to the selected route through a fixed interface. 
\textsc{Direct} uses a QA prompt, \textsc{Reduce} uses an operation-list interface, and \textsc{Execute} uses an execution-oriented interface. 
The deterministic operators used by \textsc{Reduce} are defined in Appendix~\ref{app:operator-definitions}.

\subsection{Direct QA Interface}
\label{app:direct-interface}

The \textsc{Direct} route uses a QA prompt over the serialized table. 
The prompt contains the question, the table, and formatted retained evidence. 
The evidence provides grounding context, such as resolved columns and grounded values, but it is not executed as a hard filter in this route.

For computation-related questions, \method{} may additionally provide numerical hints computed by the deterministic \textsc{Aggregate} and \textsc{Compute} operators defined in Appendix~\ref{app:operator-definitions}. 
These hints are appended to the QA prompt as auxiliary support rather than treated as mandatory answers. 
If no evidence or numerical hint is available, the route falls back to the standard QA prompt.

\subsection{Reduce Operation-List Interface}
\label{app:reduce-operation-list}

The \textsc{Reduce} route uses an operation-list interface. 
A base-model planner generates a list of operations from the question, table schema, table content, and retained evidence. 
Each operation contains three fields: \texttt{operator\_name}, \texttt{arguments}, and \texttt{explanation}. 
The \texttt{operator\_name} is selected from the predefined operator set:
\textsc{Projection}, \textsc{Filter}, \textsc{Sort}, \textsc{TopK}, \textsc{Neighbor}, \textsc{Aggregate}, and \textsc{Compute}. 
These operators cover column selection, row filtering, ordering, top-$k$ selection, relative-row lookup, aggregation, and arithmetic transformation. 
Their deterministic semantics are given in Appendix~\ref{app:operator-definitions}.

After generation, the operation list is parsed and checked by rule-based guards. 
The guards verify that referenced columns exist, operator arguments are well-formed, and the resulting plan can be executed by the table operator engine. 
The executor applies valid operators to produce a reduced table. 
For computation-related questions, it also produces a reduction context that stores outputs from \textsc{Aggregate} and \textsc{Compute}, such as counts, sums, differences, or ratios. 
This context is appended to the final QA prompt as numerical support.

\begin{table}[h]
\centering
\small
\setlength{\tabcolsep}{4pt}
\begin{tabular}{p{0.70\columnwidth}r}
\toprule
Setting & Acc. \\
\midrule
Schema-only reduction & 66.5 \\
Schema-only reduction + evidence & 68.0 \\
Best schema-only reduction & 70.1 \\
Full-table-visible reduction & 70.9 \\
\bottomrule
\end{tabular}
\caption{Reduction-only ablation on \wdata. ``Full-table-visible'' means the planner can inspect the serialized table when generating the operation list.}
\label{tab:schema-only-reduction}
\end{table}

\begin{table*}[t]
\centering
\begin{tabular}{lrrrrrr}
\toprule
Dataset & \#Q & Min Rows & Max Rows & Min Cols & Max Cols & Avg. Tokens \\
\midrule
\wdata      & 4,344 & 5  & 517       & 3 & 21 & 842.6 \\
\ldata      & 336   & 49 & 517       & 3 & 11 & 3,819.0 \\
TableBench  & 828   & 6  & 176       & 4 & 20 & 512.8 \\
\bdata      & 314   & 25 & 1,056,320 & 2 & 74 & 5,669,661.2 \\
\bottomrule
\end{tabular}
\caption{Dataset statistics. Avg. Tokens denotes the average token length of the full serialized table.}
\label{tab:dataset-stats}
\end{table*}

\subsection{Schema-Only Reduction Analysis}
\label{app:schema-only-reduction}

We analyze whether \textsc{Reduce} can rely only on the table schema when the full table is not prompt-visible. 
This is a reduction-only ablation: all examples are forced to use \textsc{Reduce}, and no routing decision is applied. 
Table~\ref{tab:schema-only-reduction} reports accuracy on WikiTQ using LLaMA3.3-70B. 
Schema-only reduction improves when evidence is added, but it remains below full-table-visible reduction. 
This supports our routing choice: when a table exceeds the model input budget, \textsc{Reduce} lacks the row- and value-level grounding needed to construct reliable operation lists.

\subsection{Evidence Use in Reduce}
\label{app:reduce-evidence-use}

Only retained evidence items are used by the \textsc{Reduce} planner. 
The fields directly used as grounding constraints are \texttt{resolved\_column}, \texttt{operator}, \texttt{value}, and \texttt{value2}. 
These fields provide column-value constraints that guide projection and filtering operations. 
Other evidence fields, such as \texttt{semantic\_unit}, \texttt{grounding\_type}, and \texttt{matched\_preview}, are used only as prompt context or diagnostics.

Validated evidence can be converted into constraints in the operation list. 
However, empty-match or uncertain evidence is treated as a soft hint rather than a mandatory filter. 
It may guide the planner towards relevant columns or values, but it is not forced into the operation list as a hard condition. 
If no evidence is retained, the evidence string is set to \texttt{None}, and the planner still attempts to generate an operation list from the question and table schema.

\subsection{Execute Interface}
\label{app:execute-interface}

The \textsc{Execute} route uses an execution-oriented interface. 
The prompt is conditioned on the question, table access, and evidence-derived constraints. 
It guides the model to generate an executable plan whose result is verified through execution. 
Validated evidence is used to ground column and value choices whenever available. 
If no evidence is retained, execution proceeds from the question and table schema or table access alone.

\subsection{Route-Specific Prompt Templates}
\label{app:route-prompts}

\info{R1-Q4, R2-Q2/3}
\fix{
Figure~\ref{fig:route-prompts} shows the prompt templates used by the
three inference routes.
Validated evidence is provided as grounding context or constraints rather
than treated as a mandatory answer.
}

\begin{figure*}[t]
\centering

\noindent
\begin{minipage}[t]{0.485\textwidth}
\vspace{0pt}
\begin{routepromptbox}{DIRECT Inference Prompt}
You are a TableQA assistant.

Answer the question using the table as the primary
evidence source. Use validated evidence and optional
computational support only as helpful context.

Return exactly:
<think>
Your reasoning
</think>
<answer>
Your final answer
</answer>

Question:
{question}

Validated Evidence:
{evidence}

Computational Support:
{computation}

Table:
{table}
\end{routepromptbox}

\centering
\smallskip
{\small (a) \readroute{} inference prompt.}
\end{minipage}%
\hfill
\begin{minipage}[t]{0.485\textwidth}
\vspace{0pt}
\begin{routepromptbox}{REDUCE Planning Prompt}
You are a TableQA decomposer.

Generate the minimal executable operation list needed
to answer the question.

Allowed operators:
PROJECTION, FILTER, NEIGHBOR, SORT, TOPK,
AGGREGATE, COMPUTE.

Use validated evidence to ground columns, values, and
conditions. Do not invent unsupported content.

Return exactly:
<think>
Brief planning rationale
</think>
<answer>
[
  {
    "operator_name": "...",
    "arguments": {...},
    "explanation": "..."
  }
]
</answer>

Question:
{question}

Validated Evidence:
{evidence}

Required Operations:
{required_operations}

Table Schema:
{schema}

Table:
{table}
\end{routepromptbox}

\centering
\smallskip
{\small (b) \reduceroute{} planning prompt.}
\end{minipage}

\vspace{0.8em}

\begin{minipage}[t]{0.98\textwidth}
\vspace{0pt}
\begin{routepromptbox}{EXECUTE Inference Prompts}
[Schema Linking]
Select only the tables and columns required by the
question. Use validated evidence and matched values
to ground columns and conditions.

Question: {question}
Table Schema: {schema}
Matched Values: {matched_values}
Validated Evidence: {evidence}

[SQL Generation]
Generate SQLite-compatible SQL that answers the
question.

Return exactly:
<reasoning>
Brief SQL rationale
</reasoning>
<result>
SQLite query
</result>

[SQL Revision]
Repair the candidate SQL using execution or checker
feedback. Correct schema, value, condition, or
operation errors.

Candidate SQL: {candidate_sql}
Checker Feedback: {feedback}

[SQL Selection]
Select the candidate SQL most likely to execute
correctly and answer the question.

Candidates: {candidate_sqls}
\end{routepromptbox}

\centering
\smallskip
{\small (c) \execroute{} schema-linking and program-generation prompts.}
\end{minipage}

\caption{
Route-specific prompt templates used by \method{}:
(a) full-table answering with \readroute{},
(b) operation planning with \reduceroute{}, and
(c) program generation with \execroute{}.
}
\label{fig:route-prompts}
\end{figure*}

\subsection{Fallback Behavior}
\label{app:interface-fallback}

The route-specific interfaces are designed to fail softly. 
If the \textsc{Reduce} planner output cannot be parsed, the parser returns an empty operation list and the upper-level pipeline applies the configured fallback. 
If no valid reduction remains after validation and repair, the route either proceeds with an empty reduction result or falls back to a head subtable when this option is enabled. 
For \textsc{Direct} and \textsc{Execute}, missing evidence does not abort inference; the route proceeds with the available question, table, and schema information.

\begin{table}[t]
\centering
\small
\resizebox{\columnwidth}{!}{
\begin{tabular}{lcccc}
\toprule
\multirow{2}{*}{Method}
& \multicolumn{2}{c}{GPT-4o-mini}
& \multicolumn{2}{c}{Qwen2.5-72B} \\
\cmidrule(lr){2-3}
\cmidrule(lr){4-5}
& $<200$ & $\geq200$ & $<200$ & $\geq200$ \\
\midrule
End-to-End QA & 56.4 & 54.2 & 36.1 & 25.0 \\
Python QA & 52.1 & 55.6 & 34.6 & 35.4 \\
Binder & 58.5 & 60.6 & 58.5 & 58.5 \\
Dater & 53.7 & 52.8 & 52.4 & 45.3 \\
RowColRetrieval & 47.6 & 50.0 & 48.7 & 50.0 \\
TableRAG & 51.4 & 50.0 & 58.3 & 58.3 \\
TALON & \second{66.8} & \second{66.7} & \second{66.2} & \second{65.0} \\
\rowcolor{green!10}
\method{} & \best{67.7} & \best{67.2} & \best{68.7} & \best{65.5} \\
\bottomrule
\end{tabular}
}
\caption{Accuracy (\%) on \ldata{} for tables below and above 200 rows.}
\label{tab:wtql-row-bucket}
\end{table}

\begin{table}[t]
\centering
\small
\setlength{\tabcolsep}{10pt}
\begin{tabular}{lccc}
\toprule
Base LLM & CoT & RoT & \method{} \\
\midrule
LLaMA3.3-70B & 38.2 & 44.8 & \textbf{48.7} \\
\bottomrule
\end{tabular}
\caption{Generalization results on TableBench. Scores are reported with ROUGE-L.}
\label{tab:tablebench}
\end{table}

\begin{table*}[t]
\centering
\small
\setlength{\tabcolsep}{4pt}
\begin{tabular}{p{0.22\textwidth}p{0.70\textwidth}}
\toprule
\textbf{Operator} & \textbf{Pseudo-code} \\
\midrule

\texttt{projection}
&
\textbf{Function} \texttt{Projection}$(T,\ columns)$: \\
& \quad verify all columns exist in $T$ \\
& \quad $T' \leftarrow T[columns]$ \\
& \quad \textbf{return} $T'$ \\

\midrule
\texttt{filter}
&
\textbf{Function} \texttt{Filter}$(T,\ conditions)$: \\
& \quad $mask \leftarrow$ evaluate each \texttt{column--operator--value} condition \\
& \quad combine masks with specified logic \\
& \quad $T' \leftarrow T[mask]$ \\
& \quad \textbf{return} $T'$ \\

\midrule
\texttt{sort}
&
\textbf{Function} \texttt{Sort}$(T,\ column,\ order)$: \\
& \quad build typed sort key from numeric, date, or textual values \\
& \quad $T' \leftarrow$ sort $T$ by $column$ in $order$ \\
& \quad \textbf{return} $T'$ \\

\midrule
\texttt{top\_k}
&
\textbf{Function} \texttt{TopK}$(T,\ k)$: \\
& \quad $T' \leftarrow$ first $k$ rows of $T$ \\
& \quad \textbf{return} $T'$ \\

\midrule
\texttt{neighbor}
&
\textbf{Function} \texttt{Neighbor}$(T,\ anchor,\ order\_column,\ direction)$: \\
& \quad sort $T$ by $order\_column$ \\
& \quad find row matching $anchor$ \\
& \quad select adjacent row according to $direction$ \\
& \quad \textbf{return} selected row as $T'$ \\

\midrule
\texttt{aggregate}
&
\textbf{Function} \texttt{Aggregate}$(T,\ group\_by,\ aggregations)$: \\
& \quad compute statistics such as count, sum, avg, min, max \\
& \quad store results in auxiliary dictionary $z$ \\
& \quad \textbf{return} $z$ \\

\midrule
\texttt{compute}
&
\textbf{Function} \texttt{Compute}$(z,\ operation,\ left,\ right,\ alias)$: \\
& \quad read operands from $z$ or constants \\
& \quad compute add, subtract, multiply, or divide \\
& \quad $z[alias] \leftarrow$ computed value \\
& \quad \textbf{return} updated $z$ \\

\bottomrule
\end{tabular}
\caption{Pseudo-code definitions of operators used in the \textsc{Reduce} route.}
\label{tab:operator-pseudocode}
\end{table*}

\section{Dataset Statistics.}
\label{dataset}
Table~\ref{tab:dataset-stats} summarizes the scale differences across benchmarks. \wdata{} and TableBench are dominated by short tables, with average serialized lengths below 1K tokens. \ldata{} has a larger average length of 3.8K tokens and at least 49 rows per table, matching the long-table setting where full-table reading remains possible but becomes less reliable. In contrast, \bdata{} contains massive scale tables, with maximum rows exceeding one million and an average serialized length of 5.7M tokens, far beyond standard LLM context windows. These statistics support our benchmark grouping into short, long, and massive table settings.

\subsection{Row-Count Bucket Analysis on WTQ-L}
\label{app:row_count_analysis}
In addition to the aggregate results in Table~\ref{tab:main}, Table~\ref{tab:wtql-row-bucket} reports accuracy below and above 200 rows for GPT-4o-mini and Qwen2.5-72B. \method{} achieves the best accuracy in all four model--bucket settings, outperforming the strongest baseline by $0.5$--$2.5$ points. These results show that its gains persist across the evaluated row-count ranges rather than being driven by one bucket.

\subsection{TableBench Generalization Results}
\label{app:tablebench}

Table~\ref{tab:tablebench} reports the supplementary TableBench result.  TableBench covers broader table-centric reasoning tasks beyond standard TableQA. 
Using LLaMA3.3-70B, \method{} improves over both CoT and RoT, indicating that the evidence-and-routing framework can generalize beyond the three main TableQA benchmarks.

\begin{table}[t]
\centering
\small
\setlength{\tabcolsep}{5.5pt}
\begin{tabular}{lccc}
\toprule
Method & \wdata{} & \ldata{} & \bdata{} \\
\midrule
CoT~\citep{COT}
    & 76.4 & 55.9 & -- \\
RoT~\citep{rot}
    & \underline{81.7} & 56.5 & -- \\
TALON~\citep{talon}
    & 71.9 & \underline{67.9} & \underline{72.6} \\
\midrule
\method{}
    & \textbf{82.5} & \textbf{69.6} & \textbf{76.1} \\
$\Delta$
    & +0.8 & +1.7 & +3.5 \\
\bottomrule
\end{tabular}
\caption{
Exact-match accuracy (\%) using GPT-5-mini.
Bold denotes the best result and underlining the strongest baseline.
$\Delta$ is the absolute improvement of \method{} over the strongest
baseline.
}
\label{tab:gpt5mini}
\end{table}

\subsection{GPT-5-mini Generalization Results}
\label{app:gpt5mini}

\info{R3-Q4}
\fix{
Table~\ref{tab:gpt5mini} reports the supplementary results using
GPT-5-mini as the base LLM.
Across \wdata{}, \ldata{}, and \bdata{}, \method{} outperforms the
strongest baseline by 0.8, 1.7, and 3.5 points, respectively.
The larger gains on \ldata{} and \bdata{} show that long and massive tables
continue to benefit from capability-aware routing under a stronger base LLM.
}

\infon{R1-W5, R2-W4, R2-C1, R2-C3}\fixn{
\subsection{Evidence Extraction Cost and End-to-End Latency}
\label{app:additional_efficiency}

\paragraph{Evidence Extraction Cost.}
We investigate whether a smaller model can reduce the overhead of evidence generation. On \ldata{}, we replace the evidence extractor with GPT-5-nano while keeping the remaining pipeline unchanged. QA accuracy decreases from 69.0\% to 68.0\%, while total token consumption increases by 3.1\%. These results suggest that using a smaller extractor does not necessarily improve the accuracy--efficiency trade-off.

\paragraph{End-to-End Latency.}
We further measure end-to-end latency on \bdata{} using the original GPT-4o-mini DeepEye-SQL configuration. The evidence-enhanced DeepEye-SQL pipeline takes 139.1 seconds per question on average, compared with 136.6 seconds for the original DeepEye-SQL pipeline, representing a 1.9\% increase. This indicates that evidence conditioning introduces modest additional latency under the evaluated setting.
}

\section{Detailed Experimental Settings}
\label{app:exp-settings}

\paragraph{Base LLMs and serving.}
For open-source models, we use a local inference backend based on \texttt{swift.llm} with HuggingFace model weights. For GPT-4o-mini, we use the OpenAI Chat Completions API. QA decoding uses a maximum generation budget of 2048 new tokens and model default input budget. All temperatures are set to $0$ unless otherwise specified.

\paragraph{Execution route configuration.}
The \textsc{Execute} route follows the official open-source implementation of DeepEye-SQL~\citep{li2025deepeye}. 
We use its staged pipeline as the program-generation backbone, including value retrieval, schema linking, SQL generation, SQL revision, and execution-aware SQL selection.

\section{Operator Definitions}
\label{app:operator-definitions}

We define a set of deterministic table operators in Table~\ref{tab:operator-pseudocode} used by the \textsc{Reduce} route. Given a filled reduction template, the operation generator produces an executable operator list. The executor then applies these operators to obtain a reduced table and auxiliary computation information. Table-reduction operators modify the current table state, while computation-support operators update auxiliary information used by the final reasoner.

\end{document}